\documentclass[10pt,twocolumn,letterpaper]{article}

\usepackage{comment}
\usepackage{booktabs}
\usepackage{tabularx}
\usepackage{caption}
\usepackage{multirow}
\usepackage[accsupp]{axessibility}  
\usepackage[pagenumbers]{cvpr} 

\definecolor{cvprblue}{rgb}{0.21,0.49,0.74}
\usepackage[pagebackref,breaklinks,colorlinks,allcolors=cvprblue]{hyperref}

\def\paperID{15691} 
\def\confName{CVPR}
\def\confYear{2025}

\title{Data Distributional Properties \\ As Inductive Bias for Systematic Generalization}

\author{Felipe del Rio\textsuperscript{1}, Alain Raymond-Saez\textsuperscript{1}, Daniel Florea\textsuperscript{1}, Rodrigo Toro Icarte\textsuperscript{1,3}, Julio Hurtado\textsuperscript{2},\\ Cristian B. Calderon\textsuperscript{3}, Alvaro Soto\textsuperscript{1,3}\\
\textsuperscript{1}Pontificia Universidad Católica de Chile, \textsuperscript{2}University of Warwick, \textsuperscript{3}CENIA\\
{\tt\small \{fidelrio,afraymon,dflorea,rntoro\}@uc.cl, julio.hurtado@warwick.ac.uk}\\
{\tt\small cristian.buc@cenia.cl, asoto@ing.puc.cl}
}

\begin{document}
\maketitle

\begin{abstract}

Deep neural networks (DNNs) struggle at systematic generalization (SG). Several studies have evaluated the possibility of promoting SG through the proposal of novel architectures, loss functions, or training methodologies. Few studies, however, have focused on the role of training data properties in promoting SG. In this work, we investigate the impact of certain data distributional properties, as inductive biases for the SG ability of a multi-modal language model. To this end, we study three different properties. First, data diversity, instantiated as an increase in the possible values a latent property in the training distribution may take. Second, burstiness, where we probabilistically restrict the number of possible values of latent factors on particular inputs during training. Third, latent intervention, where a particular latent factor is altered randomly during training. We find that all three factors significantly enhance SG, with diversity contributing an 89\% absolute increase in accuracy in the most affected property. Through a series of experiments, we test various hypotheses to understand why these properties promote SG. Finally, we find that Normalized Mutual Information (NMI) between latent attributes in the training distribution is strongly predictive of out-of-distribution generalization. We find that a mechanism by which lower NMI induces SG is in the geometry of representations. In particular, we find that NMI induces more parallelism in neural representations (i.e., input features coded in parallel neural vectors) of the model, a property related to the capacity of reasoning by analogy. Our code is available at: \href{https://github.com/fdelrio89/data-systematic}{https://github.com/fdelrio89/data-systematic}

\end{abstract}    
\section{Introduction}
\label{sec:intro}

Humans excel at adapting quickly to novel tasks.
This feature emerges from our \textit{systematic generalization} (SG) ability, that is, the flexible recombination of previous (blocks of) knowledge in ways that allow us to behave effectively under novel task demands and, more generally, help us make sense of the world \cite{fodor1975language-of-thought,marcus2003algebraic}.

Importantly, SG depends on prior knowledge or inductive biases upon which agents can rely on \cite{lake2017building}.
Indeed, inductive biases enable the representing of knowledge in the form of discrete independent concepts, which can be leveraged for flexible and rapid adaptation to new situations.
In humans, for example, visual perception is strongly biased towards shapes which aids in recognizing previously unknown objects \cite{graham1999infants,geirhos2018shapes-vs-texture}. 
This bias helps us reason independently about color, texture, or form, aiding in recognizing unknown objects regardless of their specific combination of properties. A child who has seen many horses, for instance, is likely to infer that a zebra moves similarly to a horse despite their differing colors.


In contrast to empirical observations in humans \cite{dekker2022curriculum}, SG does not emerge naturally in an i.i.d. setting in deep neural networks (DNNs) \cite{lake2017original-scan,ruis2020gscan,kim2020cogs,keysers2019cfq}. Since training data is always finite in machine learning, models need built-in assumptions to generalize to novel inputs. These assumptions, often called inductive biases \cite{goyal2022inductive}, help guide models toward solutions in situations not seen during training, and promote model weight configurations (i.e., solutions) that encourage generalization aligned with specific objectives \cite{wolpert1995no-free-lunch,baxter2000model-inductive-bias}.
Inductive biases can take different forms, such as constraints over loss functions \cite{kukavcka2017regularization}, architectures \cite{long2015fully}, pre-training \cite{devlin-etal-2019-bert}, learning \cite{lake2023human}, or the geometry of representations \cite{ravanbakhsh2017equivariance,satorras2021n,ito2022compositional}.
However, the role of specific training data distributional properties as inductive biases remains surprisingly understudied.


A notable exception to this trend is the study of data complexity. Indeed, both in terms of patterns and scale, increasing training data complexity can improve SG in language models \cite{lake2019meta-seq2seq,jiang-etal-2022-mutual,patel-etal-2022-revisiting,zhou2023data-factors,abbasi2023CLIP-compositional-gen,fang2022data-CLIP}. Similarly, within the realm of vision, augmenting training data with translations has been shown to improve learned invariances \cite{bouchacourt2021grounding}. Nevertheless, several questions remain open. First, although incrementing training data complexity does promote SG, the property (or properties) of data distribution that subtends this promotion remains obscure.
Second, despite the tremendous success and broad applicability of vision-language models \cite{li2020oscar,li2022blip,bao2022vlmo,li2023blip}, whether data training properties can boost SG in these models has not been explored. 
Third, a systematic analysis of the effect of training data properties on internal representations and how these can promote SG is also missing. In other words, we lack a functional explanation of how data properties may promote SG.





In this work, we focus on the following research questions: \textit{"Can SG be induced solely through changing properties of the training data? If so, how and why does this work?"}. To answer these questions, we focus on a Multi-modal Masked Language Modeling (MMLM) setting and test for three different properties of the data: \textit{diversity}, an increase in the cardinality of latent factors of the training distribution;  \textit{burstiness}, a limitation on the cardinality of latent factors in particular instances of the training data; and \textit{latent intervention}, a targeted intervention on the value of a latent factor. We construct datasets that alter these properties by manipulating their generative latent factors and test how a model trained on these datasets generalizes to instances of unseen combinations of latent factors. Our experiments demonstrate that manipulating these properties increases the SG capability of a model.

Finally, to understand why these properties enhance SG, we conduct thorough experiments. We study the role of the Normalized Mutual Information (NMI) between latent attributes of a dataset and find that it correlates strongly with out-of-distribution (OOD) generalization. We then turn to the effect of data distributional properties on learned representations.  Moreover, we examine the representational geometry relative to baseline models. 
A prevailing conjecture, inspired by biological neural representations, suggests that the more parallel representations remain under a particular change in a latent attribute, the better they are at SG. Our experiments reveal that NMI influences this geometry.  Thus, our contributions are summarized as follows:


\begin{itemize}
\item We are the first to study the impact of training data properties and its effect on SG in the multimodal setting.
  \item We show that increasing the number of values of a given data factor, a property of data we call \textit{diversity}, promotes SG in an MMLM task. We achieve an absolute increase in accuracy of up to 89\% in a systematic test set, solely through manipulation of the training data.
  \item Analogously, we show that probabilistically restricting the number of values of latent factors in a given input, a term we call \textit{burstiness}, also increases SG, with an absolute 15\% increase over the baseline. This increase, however, comes at the cost of decreased performance related to the restricted factor.
  \item We also show that intervening some of the latent factors in the training data, without violating systematicity, significantly increases SG, up to an absolute increase of 15\%.

  \item While it is known that lower NMI between latent factors tend to produce models with better SG performance, we find that for datasets with the same level of NMI, performance differs greatly when altering dataset diversity. This suggests a different mechanism by which diversity increases OOD generalization.  
  
  \item We find that one mechanism by which lower NMI induces better SG is by promoting the emergence of \textit{representations with more parallellism under changes in latent attributes}. To the best of our knowledge, this mechanism for SG has not been previously reported.

\end{itemize}

\section{Problem Formulation}

We train a model on a similar task to CLEVR \citep{johnson2017clevr} to correctly respond to a textural query given contextual information (in our case, images) about a group of entities contained in it. In particular, queries that will give a partial description of the image (e.g., "\textit{small red rubber sphere, large [?] metallic cylinder, etc.}")  will always resolve to predict the value of one or more particular attributes (e.g., color in this case) of a particular entity (e.g., cube). See Figure \ref{fig:model} for an example input image, query, and expected answer. A model able to perform SG will achieve this over a test set with unseen combinations of these attributes during training. This is relevant because models tend to memorize combinations of attributes (which may prove spurious) during training rather than learn each one separately.

\subsection{Task formalization}
 Let $A$ be a set of latent attributes $\{a_1, a_2, ..., a_{|A|}\}$. Each attribute $a_i$ may take $|a_i|$ possible values. Let an entity $e$ be described by particular realizations of $A$. Thus, an entity $e_i$ can be written as an $|A|$-dimensional vector.

We define a context $c$ as a lossless representation of $N_c$ entities. Let $q$ be one or more queries about the value of a particular attribute of an entity in $c$, Let $y$ be the correct answer to queries in $q$.

Altogether, we define $\mathcal{D} = \{(c_i, q_i, y_i), i=1 .. N\}$ to be a dataset of $N$ samples, where $c_i, q_i, y_i$ refer to the context, queries and answers for the $i$-th element in the dataset. The task consists in predicting the correct $a_i$, given $c_i$ and $q_i$.

\subsection{Evaluation}

To perform our evaluation, as is standard in the systematic generalization literature, we will assume two latent attributes $\{a,b\in A\}$, where $b$ is the attribute we wish to predict systematically. We will work with two data distributions; Split A: where properties $a$ and $b$ are related; and Split B, where this relation does not hold. Specifically, Split A will contain a subset of all combinations of $a$ and $b$, while Split B will contain all remaining combinations. During our experiments, we use three different sets for training and evaluation: training ($\mathcal{D}_{train}$), test in-distribution ($\mathcal{D}_{test-ID}$), and test out-of-distribution ($\mathcal{D}_{test-OOD}$). $\mathcal{D}_{train}$ and $\mathcal{D}_{test-ID}$ come from Split A, while $\mathcal{D}_{test-OOD}$ comes from Split B. During evaluation we will focus on performance on property $b$.

\subsection{Distributional Properties of Training Data}
\label{sec:distributional_properties_training_data}
We wish to study how three properties of $\mathcal{D}_{train}$ affect SG.  Therefore, we look for distributional properties of $\mathcal{D}_{train}$ that might be able to break the link between $a$ and $b$. To achieve this, we propose three properties, define them, and give intuition on why they may work to achieve this goal.

\subsubsection{Diversity}

To disrupt the link between properties $a$ and $b$, we propose to increase the total number of values $a$ may take.  We hypothesize that increasing the diversity of values of $a$ on the training set might make it harder for the model to memorize all combinations of $a$ and $b$. Thus, the model may prefer solutions that learn to detect $a$ and $b$ separately. Therefore, we propose \textit{diversity over attribute $a$} as the total cardinality of $a \in A$, that is $div(a)=|a_i|$.  Over our experiments, we modulate diversity over $a$ in a dataset by altering the number of possible values $a$ may take in the training set.

\subsubsection{Burstiness}

Another way to influence the link between $a$ and $b$ is to disrupt the learning of $a$, such that the model may not rely on $a$ to predict $b$. To instantiate this idea as a data property, we propose the use of \textit{burstiness}. \textit{Burstiness over $a$} is defined as limiting the number of possible values $a$ may take within an individual context $c$. Intuitively, this makes batches of data where only some value of $a$ will be heavily featured, while on other batches a different set of values for $a$ will dominate. That is our intuition of \textit{burstiness}. 
To modulate \textit{burstiness}, we define a probability $p_{burst}$ that a sample is bursty or not. This property is adapted for use in the MMLM setting from previous work \citep{chan2022data-properties-drives-in-context-learning}, where it was applied by limiting the number of different class labels and eliciting a meta-learning algorithm.

\subsubsection{Latent Intervention}

Finally, another way to break this link is to directly alter the value of $a$ while keeping the value of $b$ constant. This approach is reminiscent of the causal inference literature \citep{causal_inference, Peters2016}, which uses interventions over causal variables to learn interventional probability distributions, that is, the probability distribution of outcomes should a certain intervention be affected.
Thus, \textit{latent intervention of $a$} is related to altering the values of a latent attribute consistently across all entities in a context $c$. We modulate the degree of latent intervention by the amount the attribute is altered.




\section{Experimental Setup}
\label{sec:setup}
We proceed to detail how we operationalize the ideas from the previous section.

\subsection{Dataset \label{sec:dataset}}

We create a dataset resembling the structure of the CLEVR dataset \cite{johnson2017clevr} by leveraging their generative code. Attributes for this dataset are $A = \{shape, color, material, size\}$.  Our main goal is to predict the \textit{shape} attribute systematically, while excluding certain combinations of \textit{shape-color} from $\mathcal{D}_{train}$. There are three possible shapes: \{\textit{sphere}, \textit{cube}, \textit{cylinder}\}.
The context for this dataset is a 224 x 224 color image of 3 to 10 objects. Queries are defined as a textual description of the scene where the property of the object that is to be predicted is masked. Contextual information is also included in the query to make queries unambiguous. Our training set consists of 75,000 samples and the test set of 15,000.

Properties $a$ and $b$ will be color and shape, respectively. We designate a subset of shapes (\textit{cubes} and \textit{cylinders}) to introduce systematic differences. That is, some combinations of these shapes and colors will be in Split A, while the rest will be in Split B. Spheres, on the other hand, will combine to all colors in every split, so they may act as a control.



\subsection{Distributional Properties of the Training Data}
We modulate diversity by altering the number of colors available in the $\mathcal{D}_{train}$. We test the following total number of colors: $\{8, 27, 64, 125, 216\}$. These values come from dividing the color span in each channel in $n \in \{2..6\}$ equal parts. Burstiness is implemented by limiting the number of possible colors in a given image to 3. This property is modulated by altering $p_{burst}$ with the following values: $\{0.0, 0.5, 1.0\}$. Finally, we implement latent intervention by randomly altering the hue of all colors in the image via the \textit{ColorJitter} transform in PyTorch \citep{pytorch}. Our experiments use the following values: $\{0.0, 0.05, 0.1, 0.5\}$.

\subsection{Model}
\label{subsec:model}

\begin{figure}
    \centering
    \includegraphics[width=\linewidth]{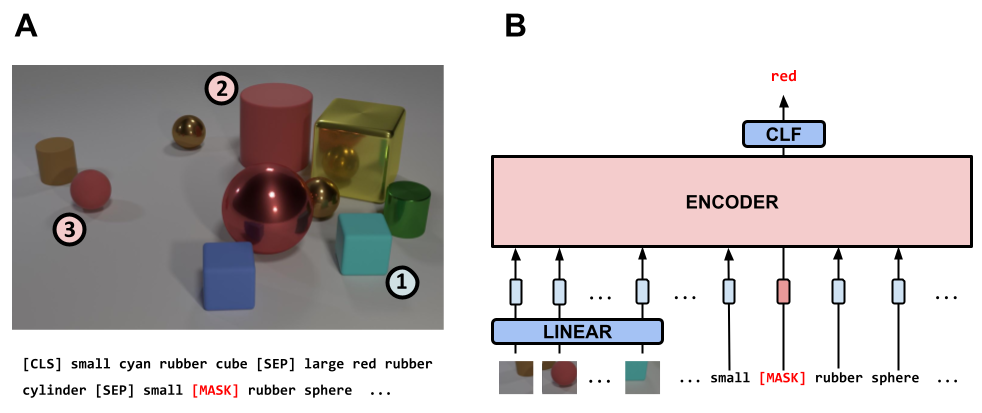}
    \caption{(A) Example input for our task. Image shows objects with different properties, while query gives a textual description of the objects while masking properties to be predicted. Sample query lists numbered objects from 1-3. (B) Diagram of the model during training. Context is an image, while the \textit{query} is text describing the scene, with some text related to properties of objects being masked. Context and query enter into a Transformer that outputs its predictions for the values of the masked properties.}
    \label{fig:model}
\end{figure}

We use a Transformer encoder \cite{vaswani2017attention} that receives both the context and query as inputs. Following standard procedure \cite{dosovitskiy2020vit}, images are divided into patches and arranged into a sequence of embeddings via a linear layer. The text is passed through an embedding layer. Learned positional and modality (image or text) embeddings are added to the input before the encoder. Both sequence embeddings are concatenated and fed into the transformer encoder (see Fig. \ref{fig:model}). 

\subsection{Evaluation}

We run each experiment for 3 seeds and report their mean and standard error. All metrics are related to performance on cubes and cylinders to correctly assess generalization in objects with systematic differences to the training set.

\section{Baseline Results on SG}
\label{sec:baseline}
\begin{figure}
    \centering    
    \includegraphics[width=0.90\linewidth]{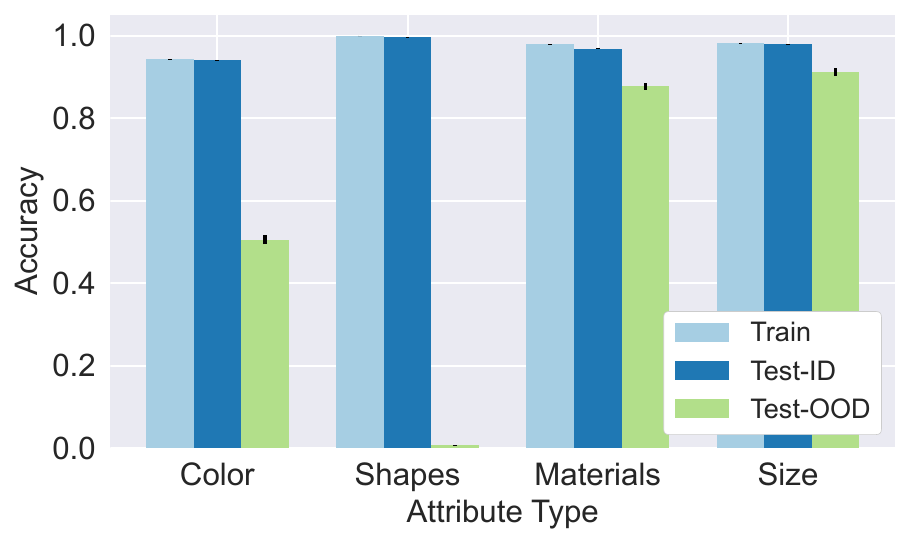}
    \caption{Accuracy of the baseline model (8 colors) for predicting different properties across data splits. While ID performance remains high, OOD performance drops sharply for \textit{shape} and \textit{color}, indicating the model learns \textit{shape-color} combinations as features, rather than achieving SG. Surprisingly, \textit{material} and \textit{size} also show a drop in OOD performance, even though the model has been exposed to all combinations of these attributes in training.}
    \label{fig:base-attributes-performance}
\end{figure}

We first analyze the behavior of a baseline model trained on 8-colors in $\mathcal{D}_{train}$, probing its ability to predict specific attributes. The results are shown in Fig. \ref{fig:base-attributes-performance}.


ID performance remains similar to that of $\mathcal{D}_{train}$ for all tasks. However, delving into performance for  $\mathcal{D}_{test-OOD}$ reveals three distinct generalization patterns. First, unexpectedly, the \textit{material} and \textit{size} show a decrease in  $\mathcal{D}_{test-OOD}$ performance, even though the model has been exposed to all combinations of these attributes during training. Second, we observe a significant decrease in performance for the \textit{color} task (44\% drop) and \textit{shape} (99\% drop). This is especially bad for the \textit{shape} task which achieves much worse than random performance (33\% to predict between three shapes in $\mathcal{D}_{test-OOD}$). These results strongly suggest that the model faces challenges when performing SG. These results confirm previous findings that indicate DNNs encounter significant performance challenges when required to generalize systematically \cite{lake2017original-scan,ruis2020gscan,kim2020cogs,keysers2019cfq}. Third, given that both shape and color are related systematically, a natural question arises: why is \textit{color} less affected than \textit{shape}? An explanation for this difference in performance between \textit{color} and \textit{shape} on $\mathcal{D}_{test-OOD}$ emerges from how image information is fed to the model. Due to the division of images into patches, the model can predict the color of a particular object solely based on the information of one patch. However, to predict the shape of an object, the model needs information from additional patches, making this ability harder to train and thus successfully achieve SG.

\begin{figure}[h]
    \centering
    \begin{subfigure}[b]{0.45\linewidth}
    \includegraphics[width=\linewidth]{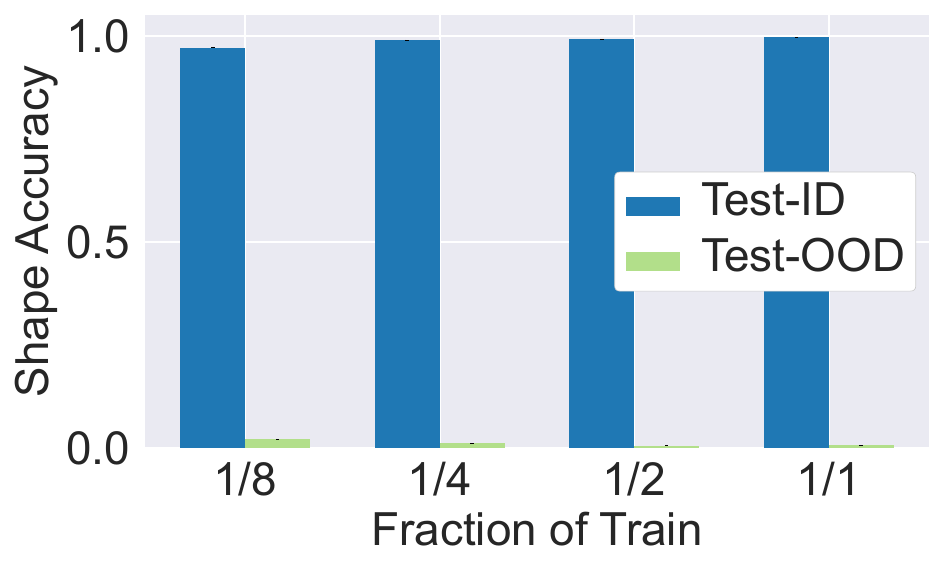}
    \caption{\centering\textit{Shape} task (8 Colors) \label{fig:baseline_less_data}}
    \end{subfigure}
    \hfill
    \begin{subfigure}[b]{0.45\linewidth}
    \includegraphics[width=\linewidth]{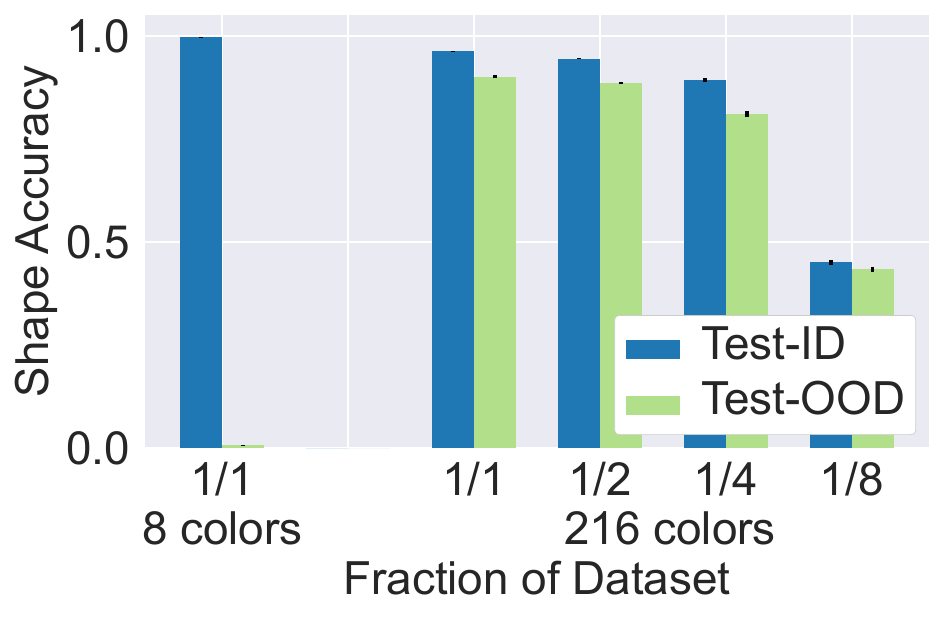}
    \caption{\centering\textit{Shape} task (216 Colors) \label{fig:sample-efficiency}}
    \end{subfigure}
    \caption{Accuracy on the \textit{shape} task with varying training data sizes for models trained on 8 colors (baseline) vs. 216 colors.
    (a) Increasing dataset size does not improve OOD performance and slightly degrades it. (b) Greater \textit{diversity} significantly enhances OOD generalization, with models trained on just a quarter of the data severely outperforming the 8-color baseline.}
    \label{fig:shapes_subset}
\end{figure}

We turn to analyzing how modulating the amount of training data affects SG. A popular way of promoting generalization is to increase the amount of training data. Therefore, we train models on different percentages of $\mathcal{D}_{train}$. Results are shown in Figure \ref{fig:baseline_less_data}. While ID roughly stays the same, increasing the amount of training data does not result in better OOD generalization. Performance degrades as dataset size increases. This suggests that merely adding more ID training data may not be a robust way of inducing a bias toward SG. Even worse, it seems to double down on whatever non-systematic bias the model is learning. We now alter the distribution of $\mathcal{D}_{train}$.

\section{Altering the Training Distribution for SG}
\label{sec:results}
We investigate the impact on SG of altering $\mathcal{D}_{train}$ by modifying properties that disrupt the link between $a$ and $b$.

\subsection{Diversity}

\begin{figure}[h]
    \centering
    \begin{subfigure}[b]{0.45\linewidth}
        \includegraphics[width=\linewidth]{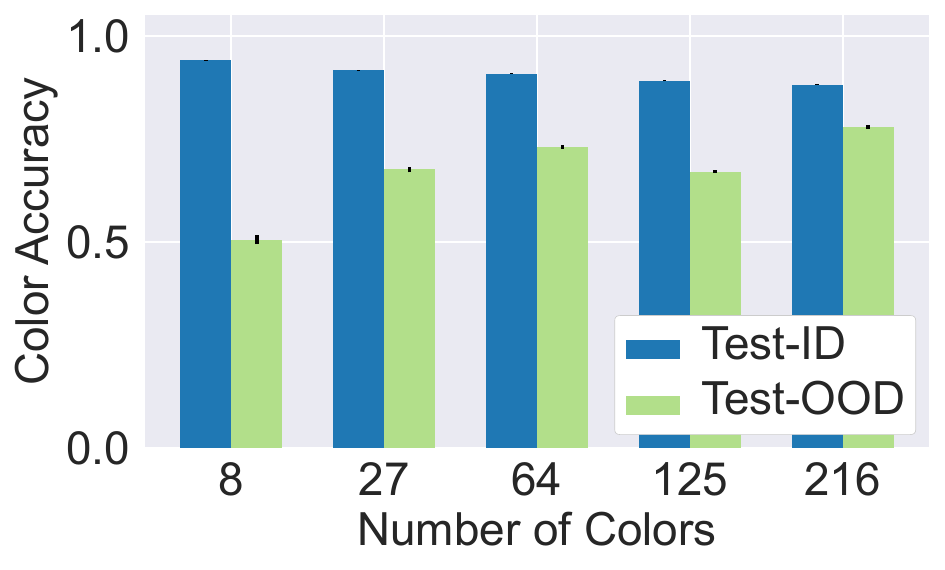}
        \caption{\textit{Color} Task}
        \label{fig:colors-acc-by-num-colors}
    \end{subfigure}
    \hspace{\fill} 
    \begin{subfigure}[b]{0.45\linewidth}
        \includegraphics[width=\linewidth]{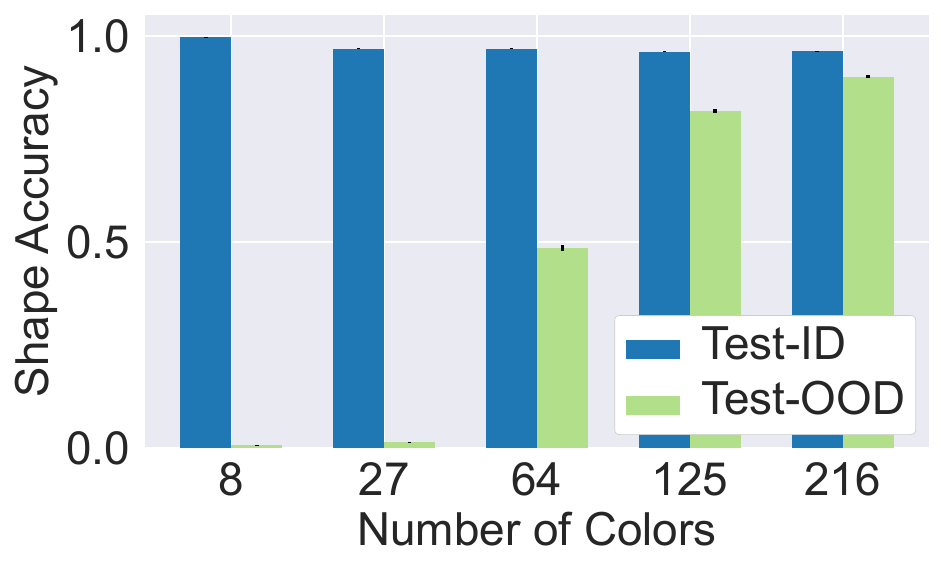}
        \caption{\textit{Shape} Task}
        \label{fig:shapes-acc-by-num-colors}
    \end{subfigure}
    
    \vspace{1em} 
    
    \begin{subfigure}[b]{0.45\linewidth}
        \includegraphics[width=\linewidth]{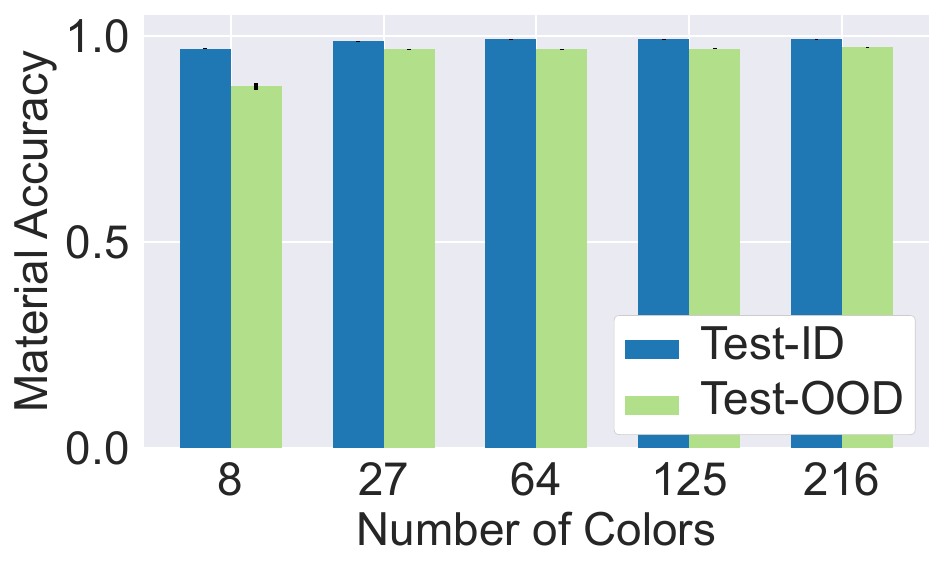}
        \caption{\textit{Material} Task}
        \label{fig:materials-acc-by-num-colors}
    \end{subfigure}
    \hspace{\fill} 
    \begin{subfigure}[b]{0.45\linewidth}
        \includegraphics[width=\linewidth]{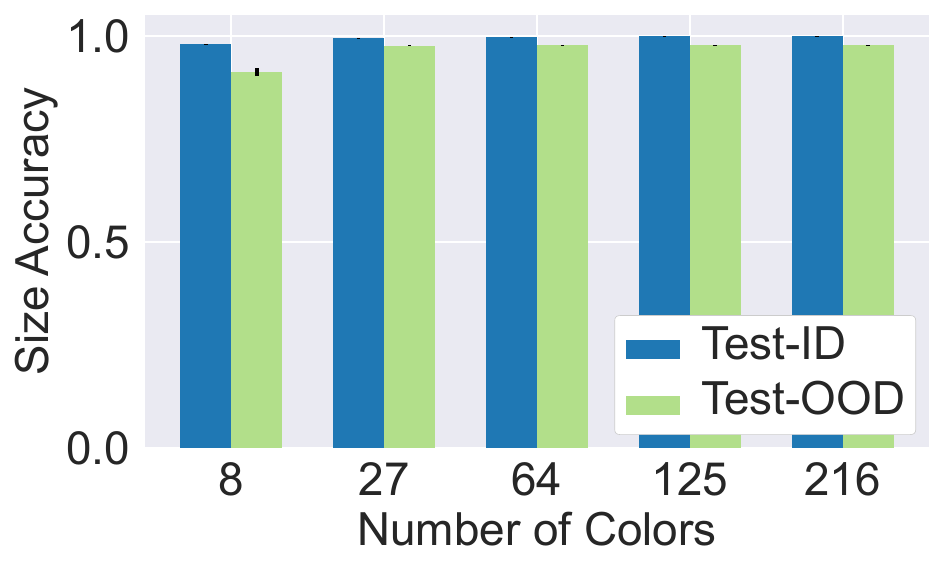}
        \caption{\textit{Size} Task}
        \label{fig:size-acc-by-num-colors}
    \end{subfigure}

    \caption{ \label{fig:acc_by_colors} Accuracy versus the number of colors in $\mathcal{D}_{train}$ for $\mathcal{D}_{test-ID}$ and $\mathcal{D}_{test-OOD}$ for the \textit{color} and \textit{shape} tasks. Performance for the \textit{shape} task increases drastically for the OOD split as we increase colors, increasing 86\% in absolute terms over the 8-color baseline. Moreover, performance in the color task also tends to increase in the OOD split, while ID only suffers slightly, even though the task becomes significantly harder. Remarkably, the \textit{material} and \textit{size} task rapidly increase their $\mathcal{D}_{test-OOD}$ performance as color increases.}

\end{figure}

To increase \textit{diversity}, we run experiments in which we increase the colors in $\mathcal{D}_{train}$ from 8 to 216. As shown in Figure \ref{fig:acc_by_colors}, greater \textit{color diversity} improves SG on the \textit{shape} task (see the monotonic increase of {\color{OliveGreen}green bars}). An absolute improvement of $89\%$, a staggering amount that reflects the difference between a model that does not generalize systematically and one that does. Surprisingly, the test-ID color prediction drops only slightly ($\sim6\%$) (Fig. \ref{fig:colors-acc-by-num-colors}), despite the increased difficulty of an increased class count. Moreover, higher color diversity reduces the performance gap between $\mathcal{D}_{test-ID}$ and $\mathcal{D}_{test-OOD}$ in the \textit{color} task (Fig. \ref{fig:colors-acc-by-num-colors}). Remarkably, OOD performance for the $material$ and $size$ tasks also increases significantly when adding colors. This suggests that the increase in diversity appears to be creating an inductive bias where at least \textit{color} is being disassociated from \textit{shape}, \textit{material}, and \textit{size}. Moreover, this increase for \textit{material} and \textit{size} occurs rapidly with just 27 colors, while for \textit{shape}, this keeps increasing until reaching 216 colors.

Finally, we assess whether increased diversity affects sample efficiency. We train models on datasets with 216 colors of $\{0.25, 0.5, 0.75\}$ the original training size and compare them to the 8-color baseline at full size. The results are shown in Figure \ref{fig:sample-efficiency}. Remarkably, even with just $25\%$ of the data with increased diversity, produces much stronger OOD generalization than the baseline, while the gap between ID and OOD remains relatively low.

\subsection{Burstiness}

\begin{figure}[h]
    \centering
    
    \begin{subfigure}[b]{0.475\linewidth}
        \captionsetup{justification=centerlast, width=0.9\linewidth}
        \includegraphics[width=\linewidth]{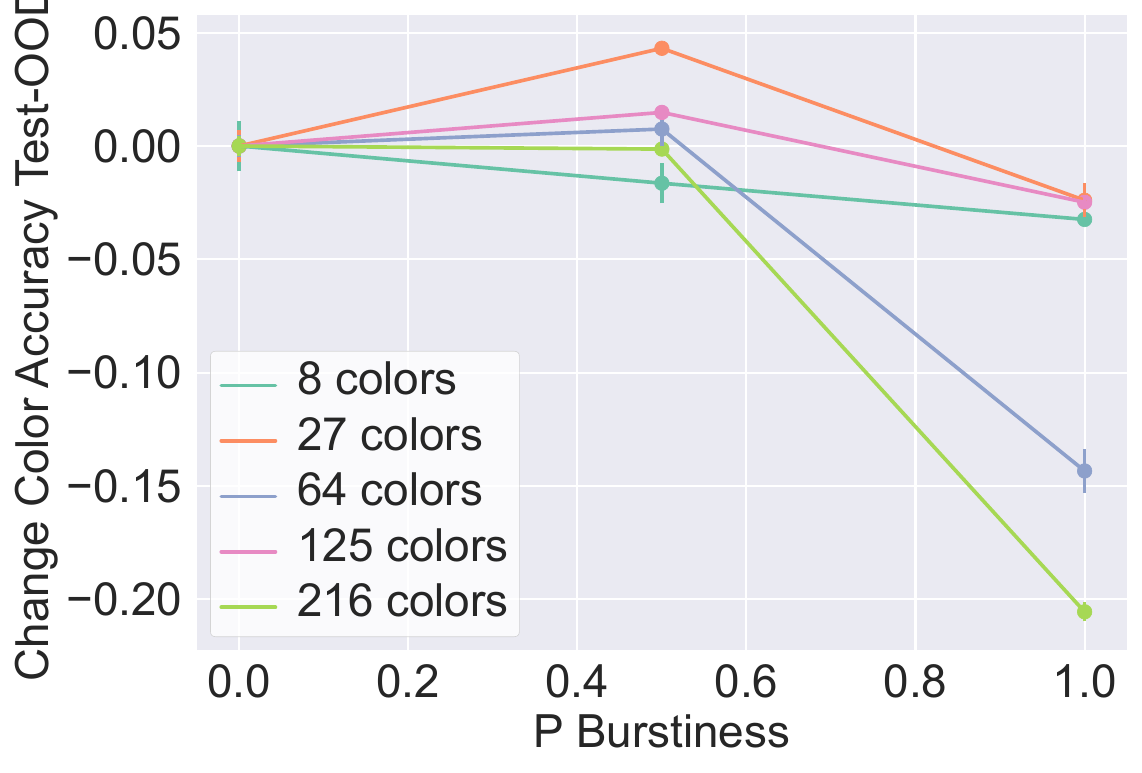}
        \caption{\textit{Color} Task}
        \label{fig:colors-common-colors-8}
    \end{subfigure}
    \hspace{\fill}
    \begin{subfigure}[b]{0.475\linewidth}
        \captionsetup{justification=centerlast, width=0.9\linewidth}
        \includegraphics[width=\linewidth]{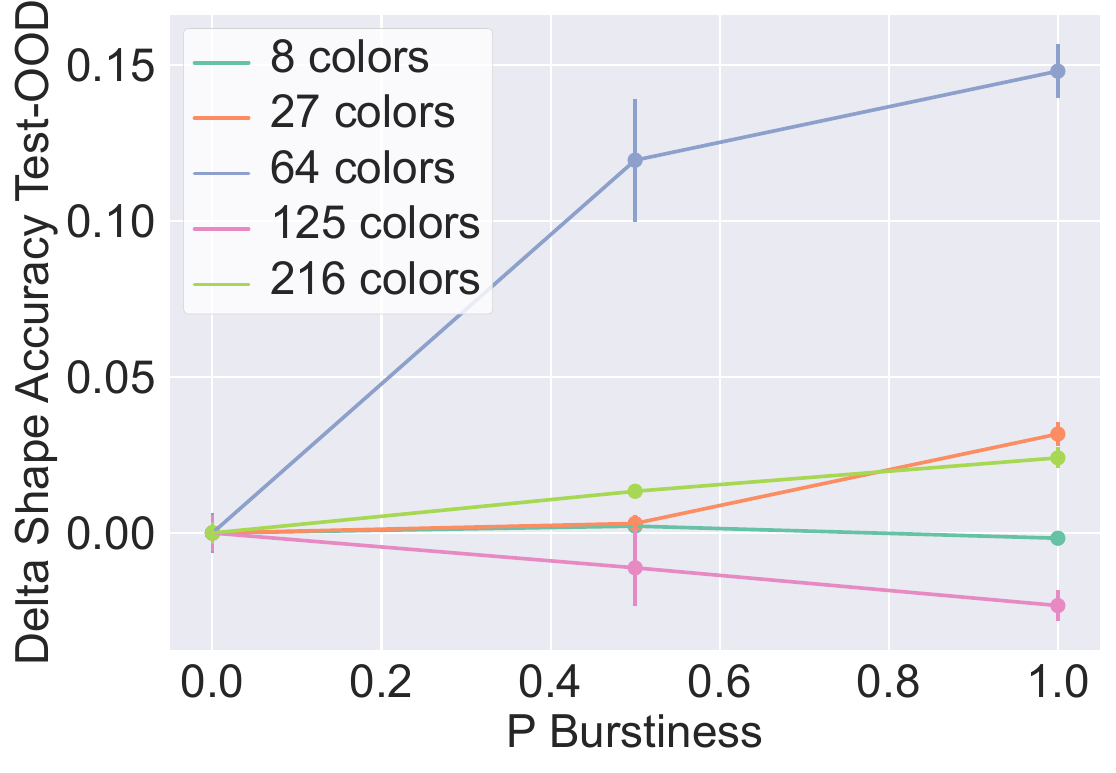}
        \caption{\textit{Shape} Task}
        \label{fig:shapes-common-colors-8}
    \end{subfigure}

    \caption{Change of accuracy with respect to no burstiness for the \textit{color} and \textit{shape} tasks in test-ID and test-OOD for different levels of \textit{burstiness} over \textit{color} for various numbers of colors. Limiting the number of colors available for each image during training allows the model to gain up to 14.8\% more accuracy over the baseline. The \textit{color} task, however, suffers up to 14.3\% decline as the \textit{color} task becomes easier to memorize.}
    \label{fig:burstiness}
\end{figure}

As was mentioned in Section \ref{sec:distributional_properties_training_data}, another way of disrupting the link between \textit{color} and \textit{shape}, is to hinder the learning of \textit{color} by limiting the number of colors contained in a given image, something we call \textit{burstiness}. Figure \ref{fig:burstiness} shows the results. As can be seen, increasing the \textit{burstiness} of the training data increases the SG capability of the model in the \textit{shape} task by up to 14.8\%. As expected,  however, the \textit{color} task is significantly affected, both in $\mathcal{D}_{test-ID}$ and $\mathcal{D}_{test-OOD}$, losing up to 14.3\% accuracy in $\mathcal{D}_{test-OOD}$ for 64 colors. This last effect is especially pronounced as we increase the number of colors contained in the dataset, which is to be expected as the \textit{color} task becomes significantly harder as the number of colors increases dramatically, while the limit of only 3 colors per image remains fixed. Finally, the effect of \textit{burstiness} on \textit{material} and \textit{size} is very small, with at most 1\% gain in $\mathcal{D}_{test-OOD}$.



\subsection{Latent Intervention}
\begin{figure}[h]
    \centering
    
    \begin{subfigure}[b]{0.475\linewidth}
        \captionsetup{justification=centerlast, width=0.9\linewidth}
        \includegraphics[width=\linewidth]{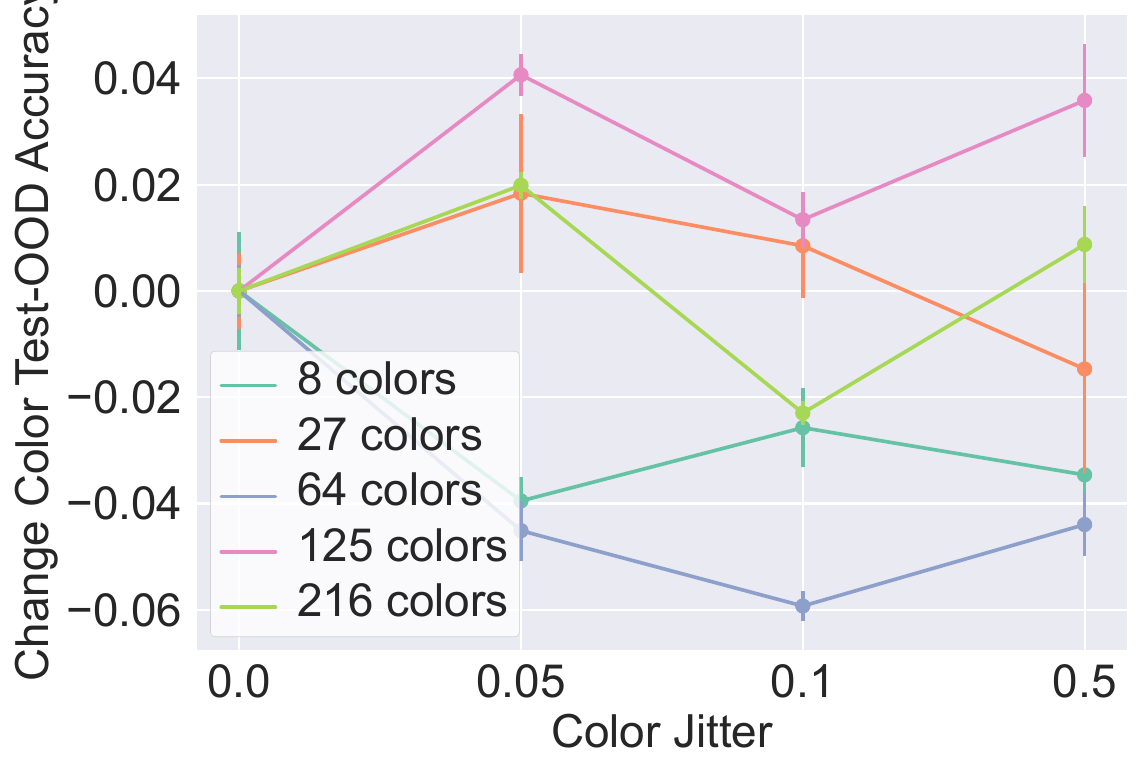}
        \caption{\textit{Color} Task}
        \label{fig:colors-common-colors-8}
    \end{subfigure}
    \hspace{\fill}
    \begin{subfigure}[b]{0.475\linewidth}
        \captionsetup{justification=centerlast, width=0.9\linewidth}
        \includegraphics[width=\linewidth]{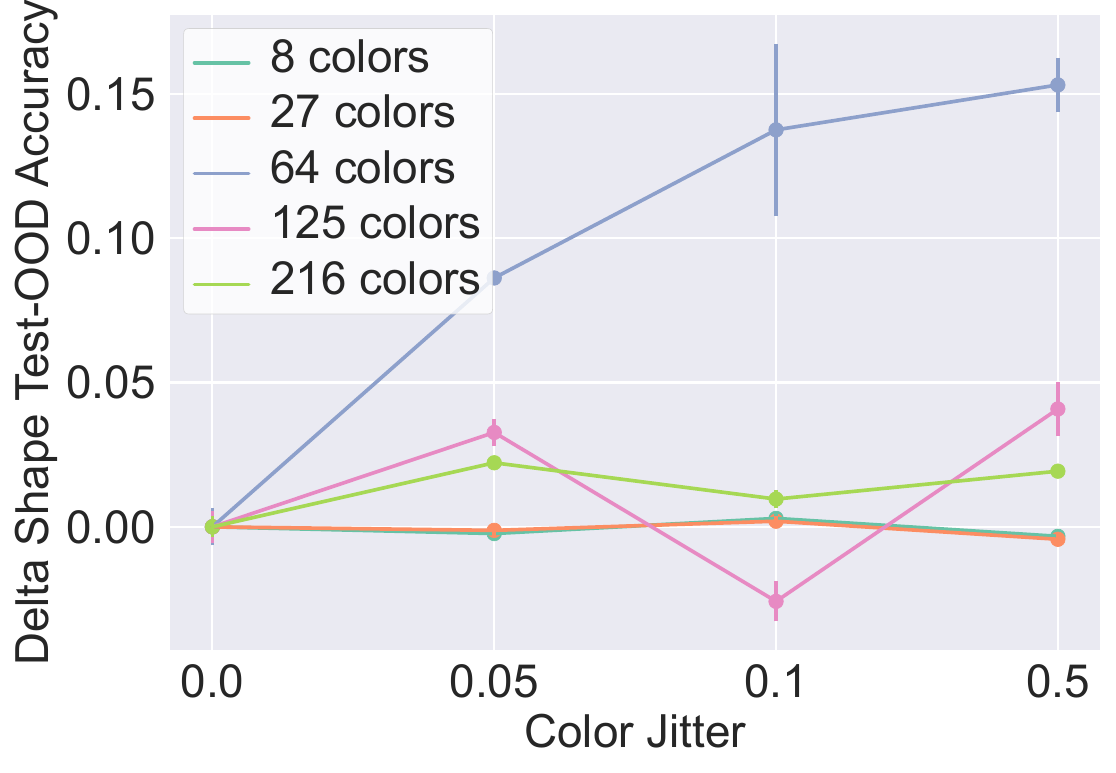}
        \caption{\textit{Shape} Task}
        \label{fig:shapes-common-colors-8}
    \end{subfigure}
    
    
    
    \caption{Change in accuracy after applying latent intervention for the \textit{color} and \textit{shape} tasks in test-ID and test-OOD for different levels of latent intervention of the \textit{color} latent attribute for various numbers of colors. X-Axis not at scale. Altering the color hue randomly during training allows the model to gain up to 15\% more OOD accuracy over baseline.}
    \label{fig:latent_intervention}
\end{figure}

Figure \ref{fig:latent_intervention} shows results for modulating \textit{latent intervention} over \textit{color}. In our experiments, we can observe that increasing the degree of the perturbation we apply to the color of the images during training increases the model performance on the \textit{shape} task by up to 15\%, a non-trivial amount while retaining OOD performance on the \textit{color} task. For the \textit{material} and \textit{size} tasks, latent intervention adds up to 2\% in $\mathcal{D}_{test-OOD}$ performance, and the effect is reduced as the number of colors increases.

\subsection{Discussion}

While all properties improve SG in the \textit{shape} task, \textit{diversity} clearly seems to have the greatest impact overall. However, we see that the other properties seem to be complementary to diversity.  This suggests a positive answer to our research question, ``\textit{Can SG be induced solely through changing properties of the training data?}".

Surprisingly, altering these properties affects not only performance on $\mathcal{D}_{test-OOD}$ for \textit{shape} and \textit{color}, but also \textit{material} and \textit{size}. This suggests that, by default, a model tends to entangle all latent features in its representation. We demonstrate that modifying the training distribution to specifically target spurious relations between latent factors can alter this behavior. We explored three approaches with varying success, addressing our second question: ``\textit{How can we modify the training distribution to induce SG?}". Finally, we turn to understanding why these properties influence SG, beginning with an analysis of model capacity.
\section{Lack of Capacity does not explain SG}

Given that \textit{diversity} has shown to improve SG performance, we would like to gain a deeper understanding of why this is the case. Our first hypothesis is the following: adding more colors to the training distribution pressures the capacity of the model, forcing it to learn simpler solutions where \textit{shape} and \textit{color} are represented independently. In other words, lack of capacity may force the model to find a more systematic solution. To test this, we test two cases: (1) We artificially restrict the capacity of a baseline Transformer model trained on 8 colors by limiting the size of the residual stream dimension ($d \in \{ 32, 64, 128 \}$). (2) We augment the capacity of the model trained on 216 colors by increasing this dimension by $\{512, 1024\}$.

\begin{figure}[h]
    \centering
    \begin{subfigure}{0.45\linewidth}
    \includegraphics[width=\linewidth]{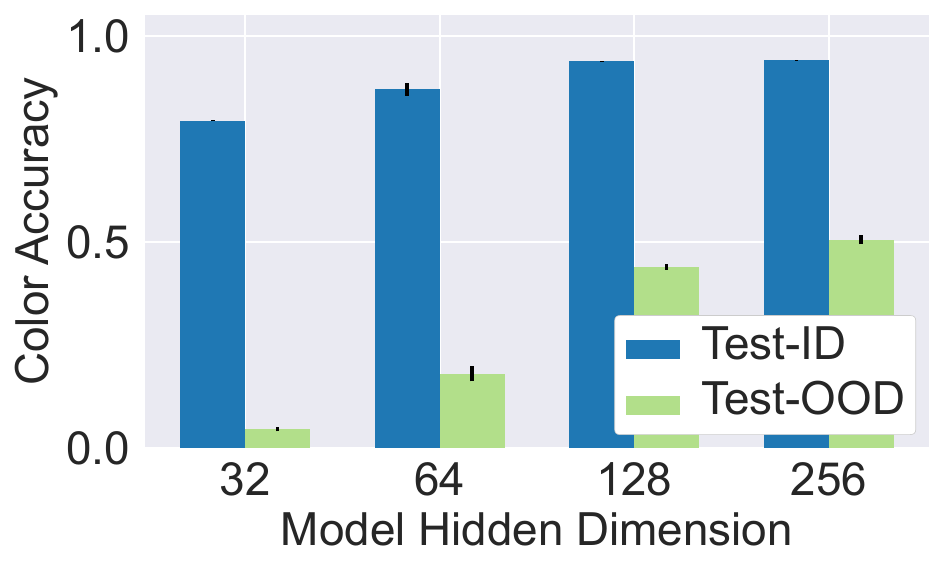}
    \caption{\textit{Color} Task}
    \end{subfigure}
    \hfill
    \begin{subfigure}{0.45\linewidth}
    \includegraphics[width=\linewidth]{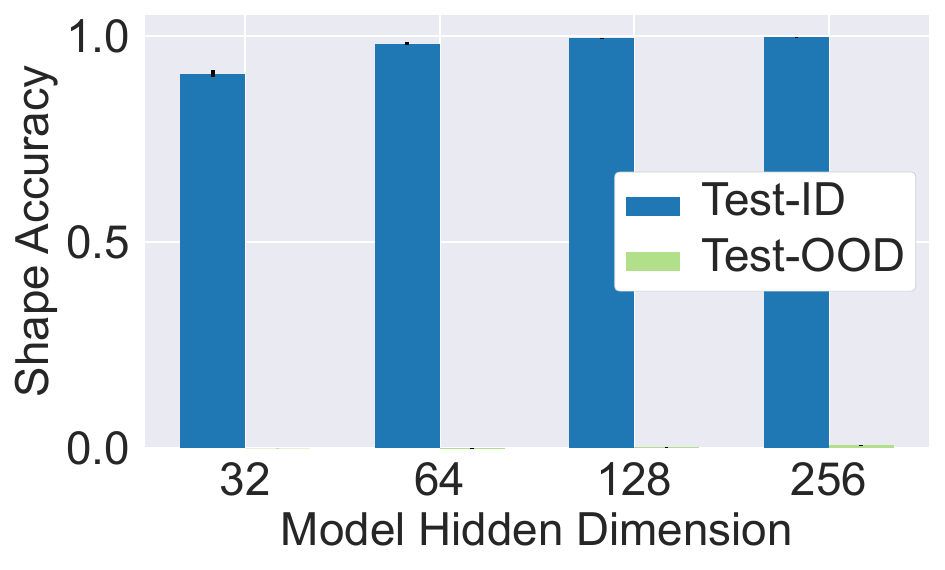}
    \caption{\textit{Shape} Task}
    \end{subfigure}

    \bigskip

    \begin{subfigure}{0.45\linewidth}
    \includegraphics[width=\linewidth]{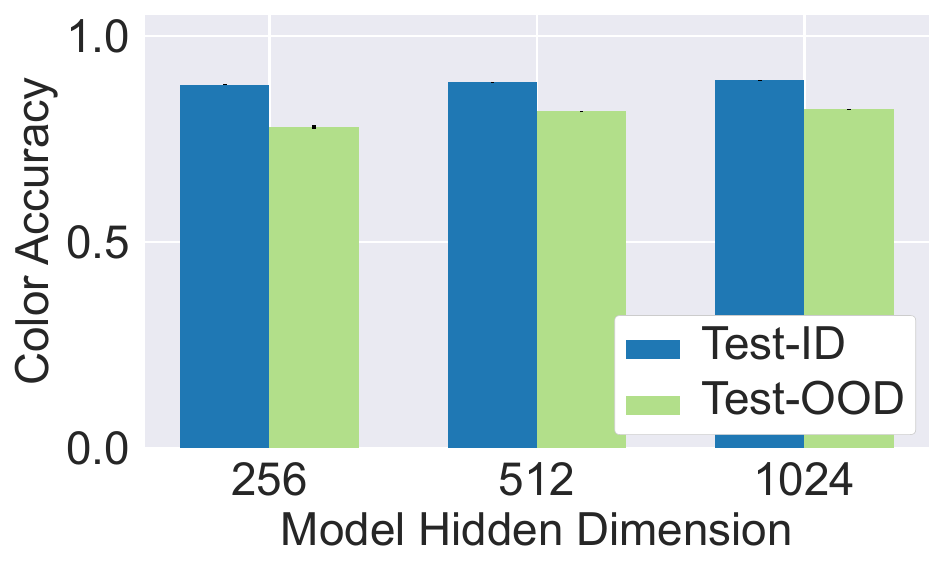}
    \caption{\textit{Color} Task}
    \end{subfigure}
    \hfill
    \begin{subfigure}{0.45\linewidth}
    \includegraphics[width=\linewidth]{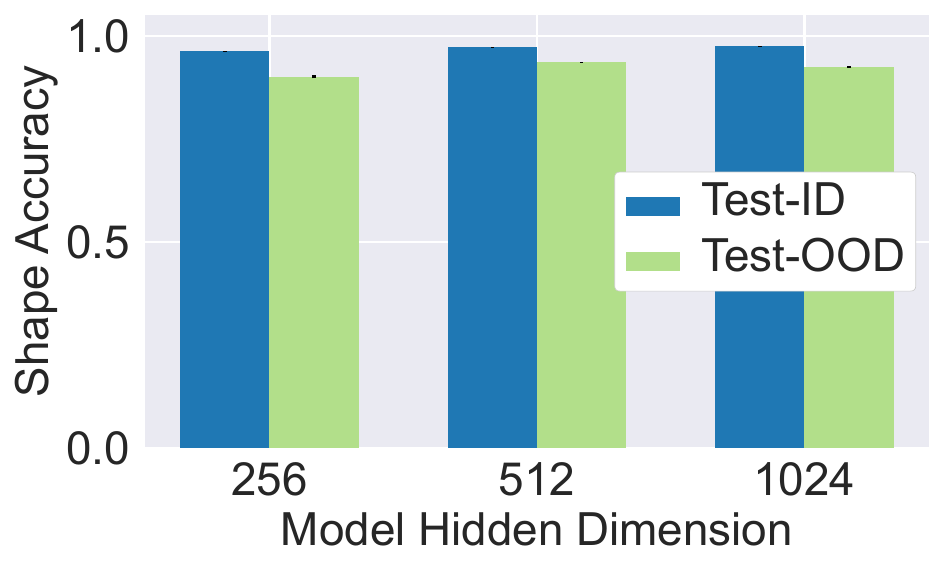}
    \caption{\textit{Shape} Task}
    \end{subfigure}
    \caption{
    {
    ID and OOD accuracy for models trained with varying hidden dimensions on a 8 (a-b) and 216 (c-d) color dataset for the  \textit{color} and \textit{shape} task. (a-b) With low \textit{diversity} (8 colors) both metrics drop as capacity decreases. (c-d) With higher \textit{diversity}, both slightly improve as capacity increases. These suggest that a capacity-bottleneck is not the cause for achieving greater SG when augmenting the number of colors in the training set.}}
    \label{fig:smaller-models}

\end{figure}

Figure \ref{fig:smaller-models} shows results for these experiments. As can be seen, performance in both $\mathcal{D}_{test-ID}$ and $\mathcal{D}_{test-OOD}$ decreases as model capacity is reduced in the first case, and the converse happens when we increase capacity. So, lack of capacity does not seem to be the mechanism by which \textit{diversity} achieves SG. We now turn to mutual information between latent attributes as a possible explanation.

\section{Mutual Information versus Diversity}

\begin{figure}
    \centering
    \includegraphics[width=0.95\linewidth]{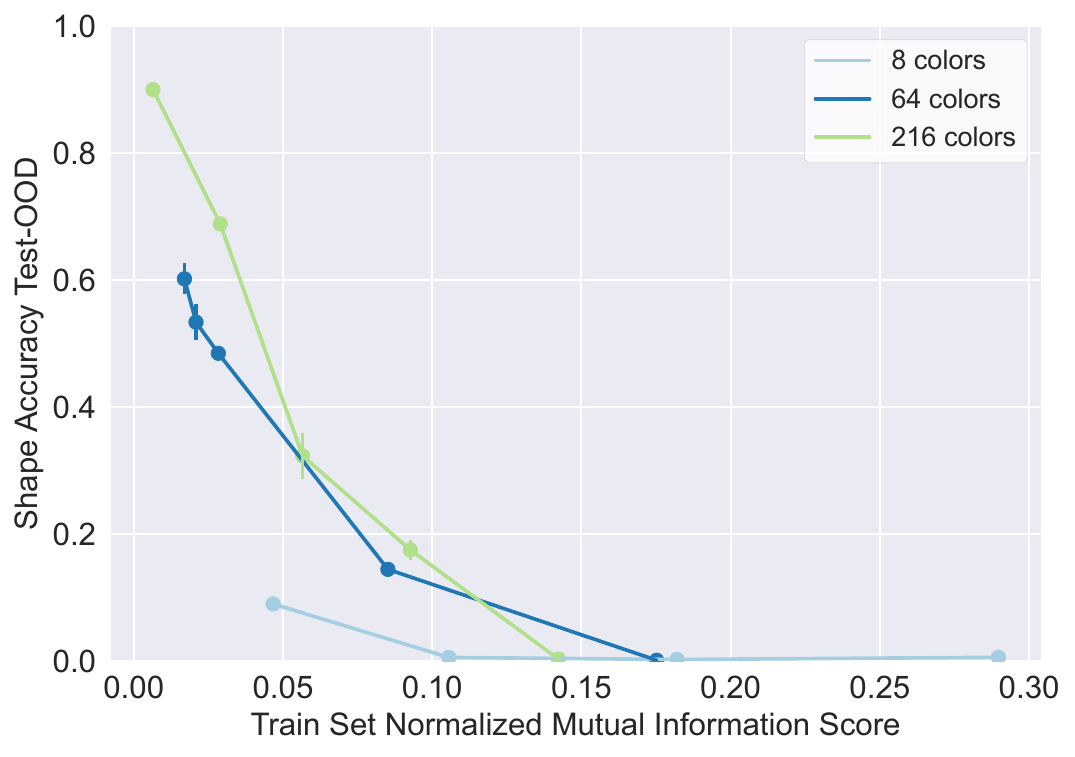}
    \caption{NMI in the training set vs OOD accuracy for the \textit{shape} task for datasets with 8, 64 and 216 colors. For lower NMI, there is greater OOD performance. However, for similar levels of NMI but differing levels of \textit{diversity}, we observe that greater \textit{diversity} increases performance. This suggests that the effect of \textit{diversity} is not only related to changing the NMI in the dataset.}
    \label{fig:nmi_vs_accuracy}
\end{figure}

Previous work \cite{abbasi2023CLIP-compositional-gen} links Normalized Mutual Information (NMI) of latent attributes and SG. Mutual information quantifies the amount of information one variable (e.g., \textit{color}) provides about another (e.g., \textit{shape}). Then, this value is normalized with the entropy of the distributions to produce the NMI. The closer NMI is to zero, the more independent these factors are. In other words, a near-zero mutual information score indicates that the dataset offers strong evidence to the model that \textit{shape} and \textit{color} should be treated as independent attributes.

Changing \textit{diversity} in a dataset impacts its NMI, so we proceed to study this effect. We seek to decouple the effect of increased \textit{diversity} from a dataset's NMI. To modulate NMI for a given level of \textit{diversity}, we produce datasets that assign colors that both \textit{cubes} and \textit{cylinders} see during training (common colors) and a set of colors that are only seen either by \textit{cubes} or by \textit{cylinders} (exclusive colors). Modulating the ratio of common colors within the dataset directly alters its NMI, while keeping the number of colors constant.

To measure the NMI, we follow previous work \cite{abbasi2023CLIP-compositional-gen} and focus on the NMI between the \textit{color} and \textit{shape} of all objects of the training dataset. It shows a Pearson correlation of -0.79. Figure \ref{fig:nmi_vs_accuracy} plots results for datasets with differing levels of \textit{diversity}. For lower NMI, there is greater OOD performance. However, for similar levels of NMI but differing levels of \textit{diversity}, we observe that greater \textit{diversity} increases performance. This suggests that \textit{diversity} influences more than just NMI, pointing to an additional mechanism. We hypothesize that this mechanism affects the geometry of representations, which we explore next.

\section{Geometry of Representations}

\subsection{Disentanglement}

The role of disentaglement in promoting SG is unclear. Some studies argue that disentanglement promotes SG by enforcing factorized representations that align with the underlying structure of the data \citep{higgins2017betavae,higgins2018definitiondisentangledrepresentations,Duan2020Unsupervised}. While other research \cite{montero2021the,xu2022compositional,schott2022visual} finds no relation between disentanglement and OOD generalization.
Using DCI metrics \citep{eastwood2018a}, we assess measures of \textit{Disentanglement} and \textit{Completeness}. \textit{Disentanglement} quantifies how many latent attributes each representation dimension predicts, ranging from 0 (equal predictive power across all attributes) to 1 (each dimension exclusively predicts a single attribute). \textit{Completeness} measures how distributed the prediction of each attribute is across dimensions, with 0 indicating equal contribution from all dimensions and 1 meaning a single dimension predicts each attribute. 

\begin{figure}[h]
    \centering
    \begin{subfigure}{0.45\linewidth}
    \includegraphics[width=\linewidth]{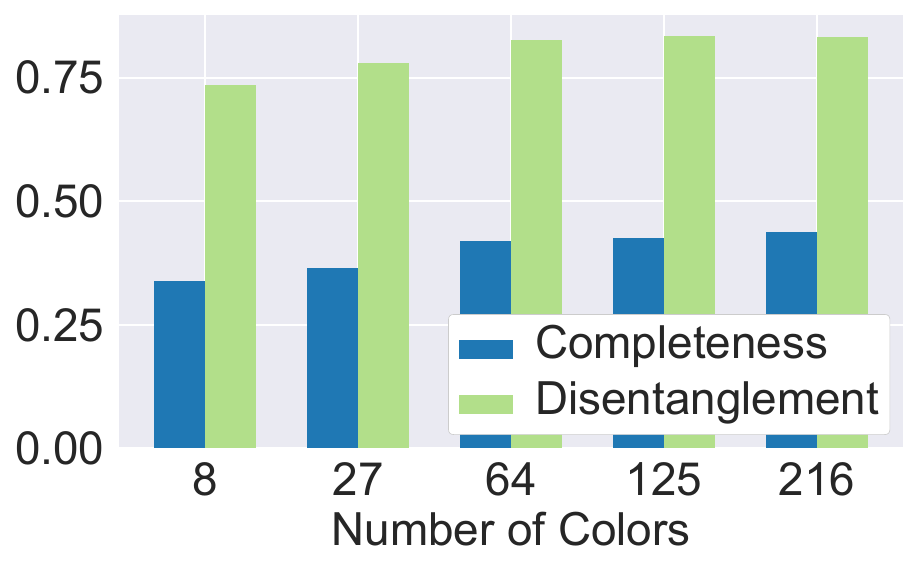}
    \caption{\textit{Shape} Task Vectors}
    \end{subfigure}
    \hfill
    \begin{subfigure}{0.45\linewidth}
    \includegraphics[width=\linewidth]{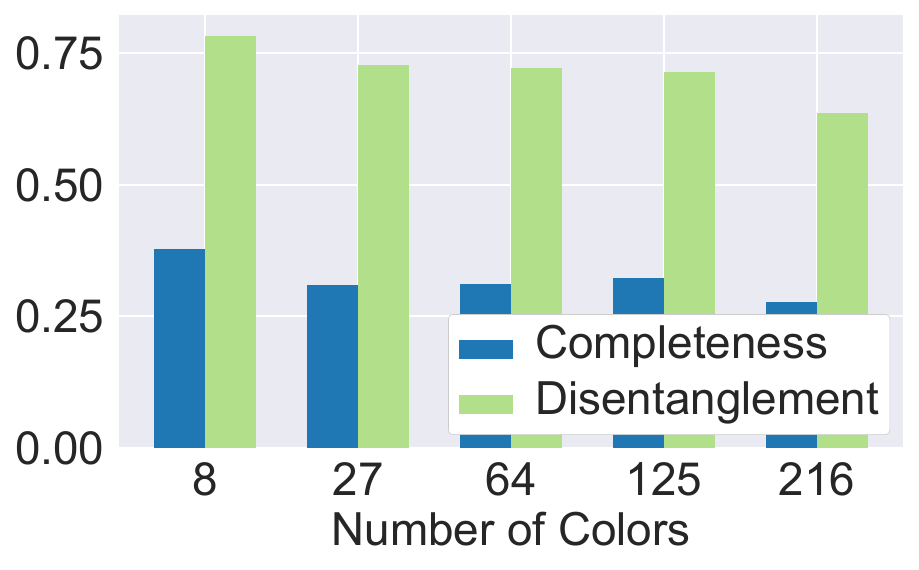}
    \caption{\textit{Color} Task Vectors}
    \end{subfigure}
    \caption{Disentanglement and Completeness for models trained on different levels of \textit{color diversity} for vectors predicting the \textit{shape} and \textit{color} tasks. Both metrics increase for the \textit{shape} task vectors, while the contrary happens for \textit{color} task vectors.}
    \label{fig:dci}

\end{figure}

Figure \ref{fig:dci} shows results for these metrics for representations for \textit{color} and \textit{shape} task, from models trained with varying color diversity. Higher diversity increases disentanglement and completeness for \textit{shape} task representations, while the converse is true for \textit{color} task representations.

We also complement our analysis by drawing from neuroscience-inspired studies highlighting how parallelism in representations enhances generalization \cite{bernardi2020gparallelism-score,ito2022parallelism-score}.

\subsection{Parallelism in Representations}

The intuition behind studying parallelism in representations is that two objects with a single different factor should have a constant direction offset in their representation, independent of other factors or context. Thus, that factor is encoded in representational space as the direction of the difference between both objects' representations.

Consider the classic example for analogy in DNNs: 
$king - man + woman = queen$ which implies:
$king - man = queen - woman = royalness$.

Here, subtracting \textit{gender} produces vectors that move in the same direction (an abstract property one might consider \textit{royalness}) for two otherwise different vectors. Encoding this property consistently in a single direction allows the model to generalize that concept through \textit{analogy}—a powerful inductive bias for solving new tasks \citep{mitchell2021abstraction}. 

We proceed to apply this same principle to the latent factors contained in our dataset. From a starting set of 1,024 images, we sample 3,500 pairs of representations from objects that share a common latent attribute. To represent each object we use the representation vector associated to the attribute we wish to study, which in this case will be either \textit{shape} or \textit{color}. We proceed to find for each of these pairs another pair of representations which represents a change in this attribute (i.e. for \textit{shape}, from a pair of red and green cubes, we look for a pair of red and green cylinders). We substract these pairs and from the resulting pair we obtain a single value called a \textit{p-score} \cite{ito2022parallelism-score}. This score is calculated by measuring the cosine of the angle between vectors in the resulting pair.  We report the mean of this value over the 3,500 pairs averaged over 5 runs.

\begin{figure}
    \centering
    \includegraphics[width=0.8\linewidth]{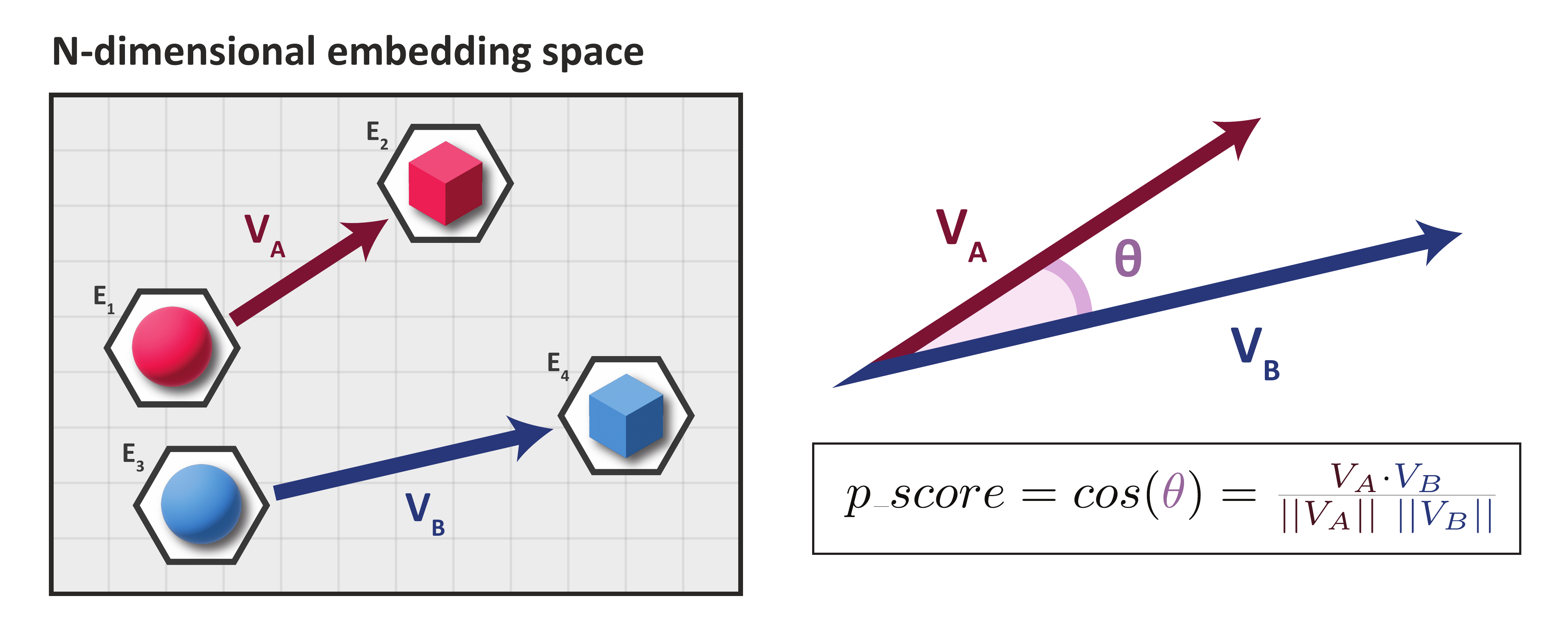}
    \caption{Computation of the \textit{p-score} metric for vectors $V_A$ and $V_B$, representing the effect of ablating features on \textit{shape} embeddings. We first subtract the embeddings to obtain vectors $V_A$ and $V_B$ and then calculate the degree of parallelism between them using cosine similarity. A high \textit{p-score} indicates strong parallelism, suggesting that the attribute (in this case, \textit{cubeness}) is encoded consistently in representational space.}
    \label{fig:pscore}
\end{figure}

\begin{figure}[h]
    \centering
    \begin{subfigure}{0.45\linewidth}
    \includegraphics[width=\linewidth]{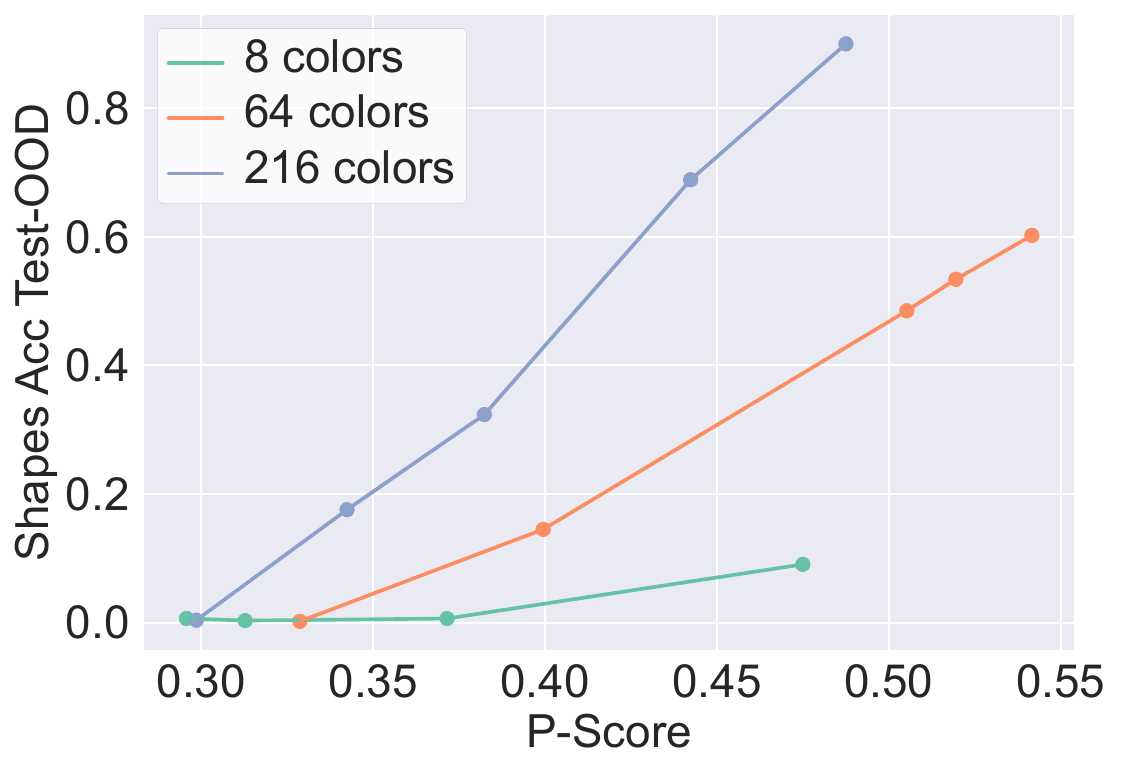}
    \caption{OOD accuracy vs. \textit{p-score} \label{fig:p-score_vs_ood}}
    \end{subfigure}
    \begin{subfigure}{0.45\linewidth}
    \includegraphics[width=\linewidth]{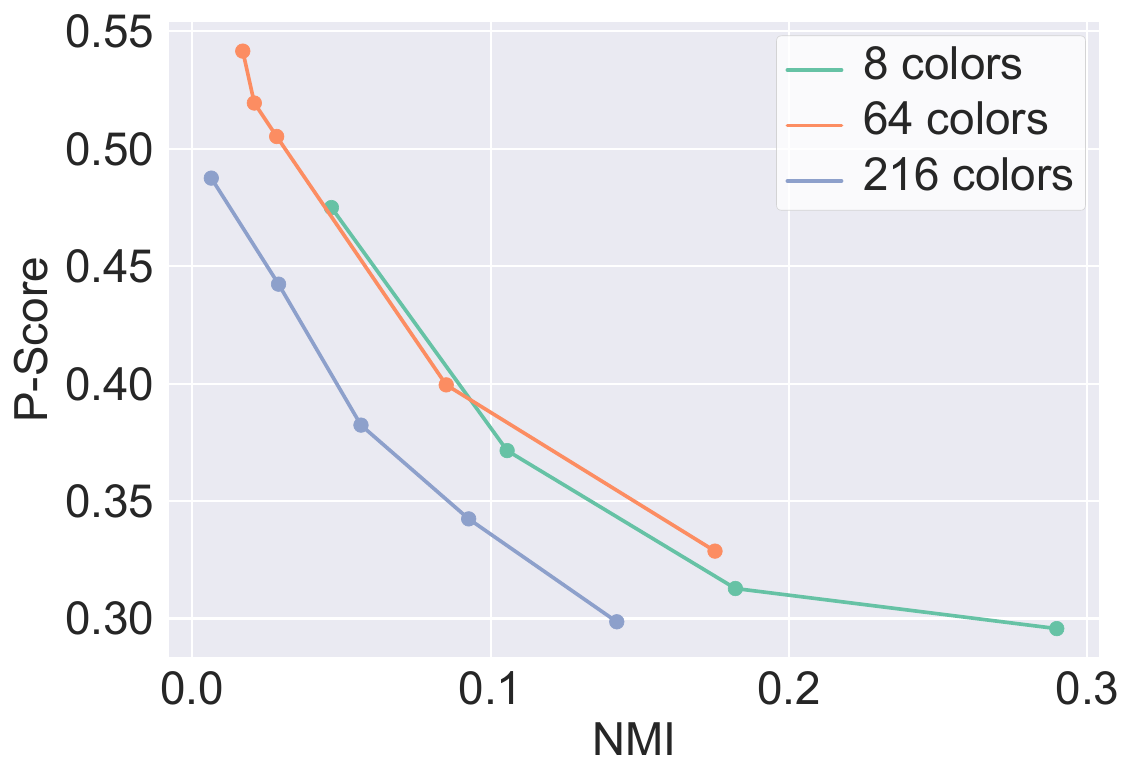}
    \caption{\textit{p-score} vs. NMI \label{fig:p-score_vs_nmi}}
    \end{subfigure}
    \caption{(a) OOD accuracy on the \textit{shape} task vs. \textit{p-score} for models trained on 8, 64, and 216 colors. Showing a trend where higher parallelism scores in the representation correlate with better performance. (b) NMI vs \textit{p-score} across models. Lower NMI datasets produce models with representations with higher \textit{p-score} suggesting a mechanism by which lower NMI aids SG.} 
    \label{fig:p-score}
\end{figure}

We relate the \textit{p-score} with performance on $\mathcal{D}_{test-OOD}$ for both the \textit{shape} and \textit{color} task for all models mentioned in Sections \ref{sec:baseline} and \ref{sec:results}. Results for experiments with $8$, $64$, and $216$ colors are shown in Figure \ref{fig:p-score_vs_ood}. Relating to the generalization capabilities in the \textit{shape} task, we see a positive trend in OOD generalization when the \textit{p-score} increases. 

To assess how significant this relationship is, we compute the Pearson correlation between \textit{p-score} and $\mathcal{D}_{test-OOD}$ performance on the \textit{shape} task (details in the supplementary material). The result, $0.73$ with very low p-value, indicates a strong link between \textit{p-score} and SG.  


We also plot the relationship between NMI and \textit{p-score}, as the positive relation both values have OOD generalization might be mediated by one or the other. Results are shown in Figure \ref{fig:p-score_vs_nmi}. A clear relationship between NMI and \textit{p-score} appears: lower NMI datasets are related to higher \textit{p-scores} in representations. Given that NMI is a stable property of the dataset, this suggests that lower NMI datasets tend to produce models with representations with higher \textit{p-scores}. This suggests a novel mechanism by which lower NMI datasets produce inductive biases during learning that allow models to achieve greater OOD generalization. These results provide evidence that the \textit{dataset distribution influences the representational space geometry of the model} biasing it towards OOD generalization.

\section{Related Work}

\paragraph{Systematic Generalization.}
Many works have addressed the problem of SG, proposing a variety of strategies. These include data augmentation methods \cite{andreas2020geca,akyureklearning,yang2022subs,akyurek-andreas-2023-lexsym}, training methods \cite{lake2019meta-seq2seq,conklin2021meta,lake2023human}, adjustments to the loss function \cite{heinze2020gscan-auxiliary}, architectural changes \cite{dessi2019cnns,bahdanau2019sys-gen-what-required,dubois2020location,kuo2020gscan-recursive,gao2020gscan-gnn,ontanon2021transformer-sys-gen,qiu2021gscan-solved} and neuro-symbolic approaches  \cite{liu2020compositional,nye2020lcomp-program-synth}. Currently, state-of-the-art methods are based on enhanced LLMs \cite{drozdov2022compositional, qiu2022improving}.

\paragraph{Data-driven Inductive Biases.} Previous work has investigated the invariances \cite{bouchacourt2021grounding}, inductive biases such as shape bias \cite{geirhos2018shapes-vs-texture}, and robustness \cite{fang2022data-CLIP} acquired by deep learning models when trained on large-scale datasets.
\citet{prystawski2024think} shows the need for local structure in the training data for models to learn probabilistic reasoning abilities.

Other studies have examined the data properties that enable models to generalize OOD. \citet{chan2022data-properties-drives-in-context-learning} show that burstiness, long-tailed distributions and label mappings can foster in-context learning in Transformers. By contrast, our work focuses specifically on data properties that support SG. For SG, studies have shown that increasing the number of primitives or input length can enhance SG \cite{patel2022revisiting,zhou2023data-factors,rahimi2024d3}. However, their focus is on generalization in the text domain, while we focus on images, examining how different latent factors within the dataset interact to drive SG. 

Other work on generalization in the visual domain \citep{abbasi2023CLIP-compositional-gen} has investigated SG in CLIP models, showing that statistical independence between object-attribute pairs and disentangled representations enhances SG. In contrast, we analyze the effects of explicit generative factors on SG and the mechanisms by which it is enforced.

\paragraph{Linear Representation Hypothesis} The parallelism of neural network representations is closely related to the \textit{linear representation hypothesis}. This hypothesis suggests that neural network activations can be understood in terms of distinct features, which correspond to specific directions in activation space. Prior research has provided evidence for this hypothesis in both categorical \cite{mikolov2013linguistic,olah2020zoom,elhage2022toy-models,marks2024geometry-truth,tigges2024language,AllenZhu-icml2024-tutorial,gould2024successor,park2024linear} and numerical features \cite{gurnee2023lan-space-time,heinzerling-inui-2024-monotonic}. Furthermore, it has been proposed that learning linear representations can enhance OOD generalization by reducing the nonlinear entanglement or contamination of different features \cite{zhang2024feature-contamination}.




\section{Conclusion}

We propose a different way of shaping inductive biases for SG, focusing on \textit{altering the properties of training data} over other techniques. We show three properties of the training data that, when altered, significantly boost SG: \textit{diversity}, \textit{burstiness}, and \textit{latent intervention}. \textit{Diversity} has the strongest impact, improving SG by up to an absolute 89\% over baseline, while the other factors complement it.

We also study why these are effective. We show that limiting model capacity is not a factor for SG, while increased diversity and lower NMI between latent attributes in the dataset seem to explain most of the gains in SG. Crucially, we test for a property of brain-like representations that measures parallelism in representations under changes in a latent attribute. We find that datasets with lower NMI tend to induce more of this parallelism, a property that allows models to extrapolate by analogy. This is, to the best of our knowledge, an unreported mechanism by which dataset distribution induces SG. 


\section*{Acknowledgements}

This work was supported by FONDECYT under Grants 1221425 and 11230762;
and National Center for Artificial Intelligence (CENIA) under Grant FB210017 Basal ANID.
We gratefully acknowledge the support of the AMD University Program (AUP), whose computing resources enabled the experiments conducted in this work.

{
    \small
    \bibliographystyle{ieeenat_fullname}
    \bibliography{main}

\begin{thebibliography}{80}
\providecommand{\natexlab}[1]{#1}
\providecommand{\url}[1]{\texttt{#1}}
\expandafter\ifx\csname urlstyle\endcsname\relax
  \providecommand{\doi}[1]{doi: #1}\else
  \providecommand{\doi}{doi: \begingroup \urlstyle{rm}\Url}\fi

\bibitem[Abbasi et~al.(2024)Abbasi, Rohban, and Baghshah]{abbasi2023CLIP-compositional-gen}
Reza Abbasi, Mohammad~Hossein Rohban, and Mahdieh~Soleymani Baghshah.
\newblock Deciphering the role of representation disentanglement: Investigating compositional generalization in clip models, 2024.

\bibitem[Akyurek and Andreas(2023)]{akyurek-andreas-2023-lexsym}
Ekin Akyurek and Jacob Andreas.
\newblock {L}ex{S}ym: Compositionality as lexical symmetry.
\newblock In \emph{Proceedings of the 61st Annual Meeting of the Association for Computational Linguistics (Volume 1: Long Papers)}, pages 639--657, Toronto, Canada, 2023. Association for Computational Linguistics.

\bibitem[Aky{\"u}rek et~al.()Aky{\"u}rek, Aky{\"u}rek, and Andreas]{akyureklearning}
Ekin Aky{\"u}rek, Afra~Feyza Aky{\"u}rek, and Jacob Andreas.
\newblock Learning to recombine and resample data for compositional generalization.
\newblock In \emph{International Conference on Learning Representations}.

\bibitem[{Allen-Zhu}(2024)]{AllenZhu-icml2024-tutorial}
Zeyuan {Allen-Zhu}.
\newblock {ICML 2024 Tutorial: Physics of Language Models}, 2024.
\newblock Project page: \url{https://physics.allen-zhu.com/}.

\bibitem[Andreas(2020)]{andreas2020geca}
Jacob Andreas.
\newblock Good-enough compositional data augmentation.
\newblock In \emph{Proceedings of the 58th Annual Meeting of the Association for Computational Linguistics}, pages 7556--7566, 2020.

\bibitem[Bahdanau et~al.(2019)Bahdanau, Murty, Noukhovitch, Nguyen, Vries, and Courville]{bahdanau2019sys-gen-what-required}
Dzmitry Bahdanau, Shikhar Murty, Michael Noukhovitch, Thien~Huu Nguyen, Harm~de Vries, and Aaron Courville.
\newblock Systematic generalization: What is required and can it be learned?
\newblock In \emph{International Conference on Learning Representations}, 2019.

\bibitem[Bao et~al.(2022)Bao, Wang, Dong, Liu, Mohammed, Aggarwal, Som, Piao, and Wei]{bao2022vlmo}
Hangbo Bao, Wenhui Wang, Li Dong, Qiang Liu, Owais~Khan Mohammed, Kriti Aggarwal, Subhojit Som, Songhao Piao, and Furu Wei.
\newblock Vlmo: Unified vision-language pre-training with mixture-of-modality-experts.
\newblock \emph{Advances in Neural Information Processing Systems}, 35:\penalty0 32897--32912, 2022.

\bibitem[Baxter(2000)]{baxter2000model-inductive-bias}
Jonathan Baxter.
\newblock A model of inductive bias learning.
\newblock \emph{Journal of artificial intelligence research}, 12:\penalty0 149--198, 2000.

\bibitem[Bernardi et~al.(2020)Bernardi, Benna, Rigotti, Munuera, Fusi, and Salzman]{bernardi2020gparallelism-score}
Silvia Bernardi, Marcus~K Benna, Mattia Rigotti, J{\'e}r{\^o}me Munuera, Stefano Fusi, and C~Daniel Salzman.
\newblock The geometry of abstraction in the hippocampus and prefrontal cortex.
\newblock \emph{Cell}, 183\penalty0 (4):\penalty0 954--967, 2020.

\bibitem[Bouchacourt et~al.(2021)Bouchacourt, Ibrahim, and Morcos]{bouchacourt2021grounding}
Diane Bouchacourt, Mark Ibrahim, and Ari Morcos.
\newblock Grounding inductive biases in natural images: invariance stems from variations in data.
\newblock \emph{Advances in Neural Information Processing Systems}, 34:\penalty0 19566--19579, 2021.

\bibitem[Chan et~al.(2022)Chan, Santoro, Lampinen, Wang, Singh, Richemond, McClelland, and Hill]{chan2022data-properties-drives-in-context-learning}
Stephanie Chan, Adam Santoro, Andrew Lampinen, Jane Wang, Aaditya Singh, Pierre Richemond, James McClelland, and Felix Hill.
\newblock Data distributional properties drive emergent in-context learning in transformers.
\newblock \emph{Advances in Neural Information Processing Systems}, 35:\penalty0 18878--18891, 2022.

\bibitem[Conklin et~al.(2021)Conklin, Wang, Smith, and Titov]{conklin2021meta}
Henry Conklin, Bailin Wang, Kenny Smith, and Ivan Titov.
\newblock Meta-learning to compositionally generalize.
\newblock In \emph{Proceedings of the 59th Annual Meeting of the Association for Computational Linguistics and the 11th International Joint Conference on Natural Language Processing (Volume 1: Long Papers)}, pages 3322--3335, 2021.

\bibitem[Dekker et~al.(2022)Dekker, Otto, and Summerfield]{dekker2022curriculum}
Ronald~B Dekker, Fabian Otto, and Christopher Summerfield.
\newblock Curriculum learning for human compositional generalization.
\newblock \emph{Proceedings of the National Academy of Sciences}, 119\penalty0 (41):\penalty0 e2205582119, 2022.

\bibitem[Dess{\`\i} and Baroni(2019)]{dessi2019cnns}
Roberto Dess{\`\i} and Marco Baroni.
\newblock Cnns found to jump around more skillfully than rnns: Compositional generalization in seq2seq convolutional networks.
\newblock In \emph{Proceedings of the 57th Annual Meeting of the Association for Computational Linguistics}, pages 3919--3923, 2019.

\bibitem[Devlin et~al.(2019)Devlin, Chang, Lee, and Toutanova]{devlin-etal-2019-bert}
Jacob Devlin, Ming-Wei Chang, Kenton Lee, and Kristina Toutanova.
\newblock {BERT}: Pre-training of deep bidirectional transformers for language understanding.
\newblock In \emph{Proceedings of the 2019 Conference of the North {A}merican Chapter of the Association for Computational Linguistics: Human Language Technologies, Volume 1 (Long and Short Papers)}, pages 4171--4186, Minneapolis, Minnesota, 2019. Association for Computational Linguistics.

\bibitem[Dosovitskiy et~al.(2020)Dosovitskiy, Beyer, Kolesnikov, Weissenborn, Zhai, Unterthiner, Dehghani, Minderer, Heigold, Gelly, et~al.]{dosovitskiy2020vit}
Alexey Dosovitskiy, Lucas Beyer, Alexander Kolesnikov, Dirk Weissenborn, Xiaohua Zhai, Thomas Unterthiner, Mostafa Dehghani, Matthias Minderer, Georg Heigold, Sylvain Gelly, et~al.
\newblock An image is worth 16x16 words: Transformers for image recognition at scale.
\newblock In \emph{International Conference on Learning Representations}, 2020.

\bibitem[Drozdov et~al.()Drozdov, Sch{\"a}rli, Aky{\"u}rek, Scales, Song, Chen, Bousquet, and Zhou]{drozdov2022compositional}
Andrew Drozdov, Nathanael Sch{\"a}rli, Ekin Aky{\"u}rek, Nathan Scales, Xinying Song, Xinyun Chen, Olivier Bousquet, and Denny Zhou.
\newblock Compositional semantic parsing with large language models.
\newblock In \emph{The Eleventh International Conference on Learning Representations}.

\bibitem[Duan et~al.(2020)Duan, Matthey, Saraiva, Watters, Burgess, Lerchner, and Higgins]{Duan2020Unsupervised}
Sunny Duan, Loic Matthey, Andre Saraiva, Nick Watters, Chris Burgess, Alexander Lerchner, and Irina Higgins.
\newblock Unsupervised model selection for variational disentangled representation learning.
\newblock In \emph{International Conference on Learning Representations}, 2020.

\bibitem[Dubois et~al.(2020)Dubois, Dagan, Hupkes, and Bruni]{dubois2020location}
Yann Dubois, Gautier Dagan, Dieuwke Hupkes, and Elia Bruni.
\newblock Location attention for extrapolation to longer sequences.
\newblock In \emph{Proceedings of the 58th Annual Meeting of the Association for Computational Linguistics}, pages 403--413, 2020.

\bibitem[Eastwood and Williams(2018)]{eastwood2018a}
Cian Eastwood and Christopher K.~I. Williams.
\newblock A framework for the quantitative evaluation of disentangled representations.
\newblock In \emph{International Conference on Learning Representations}, 2018.

\bibitem[Elhage et~al.(2022)Elhage, Hume, Olsson, Schiefer, Henighan, Kravec, Hatfield-Dodds, Lasenby, Drain, Chen, et~al.]{elhage2022toy-models}
Nelson Elhage, Tristan Hume, Catherine Olsson, Nicholas Schiefer, Tom Henighan, Shauna Kravec, Zac Hatfield-Dodds, Robert Lasenby, Dawn Drain, Carol Chen, et~al.
\newblock Toy models of superposition.
\newblock \emph{arXiv preprint arXiv:2209.10652}, 2022.

\bibitem[Fang et~al.(2022)Fang, Ilharco, Wortsman, Wan, Shankar, Dave, and Schmidt]{fang2022data-CLIP}
Alex Fang, Gabriel Ilharco, Mitchell Wortsman, Yuhao Wan, Vaishaal Shankar, Achal Dave, and Ludwig Schmidt.
\newblock Data {{Determines Distributional Robustness}} in {{Contrastive Language Image Pre-training}} ({{CLIP}}).
\newblock In \emph{Proceedings of the 39th {{International Conference}} on {{Machine Learning}}}, pages 6216--6234. PMLR, 2022.

\bibitem[Fodor(1975)]{fodor1975language-of-thought}
Jerry~A Fodor.
\newblock \emph{The language of thought}.
\newblock Harvard university press, 1975.

\bibitem[Gao et~al.(2020)Gao, Huang, and Mooney]{gao2020gscan-gnn}
Tong Gao, Qi Huang, and Raymond Mooney.
\newblock Systematic generalization on g{SCAN} with language conditioned embedding.
\newblock In \emph{Proceedings of the 1st Conference of the Asia-Pacific Chapter of the Association for Computational Linguistics and the 10th International Joint Conference on Natural Language Processing}, pages 491--503, Suzhou, China, 2020. Association for Computational Linguistics.

\bibitem[Geirhos et~al.(2019)Geirhos, Rubisch, Michaelis, Bethge, Wichmann, and Brendel]{geirhos2018shapes-vs-texture}
Robert Geirhos, Patricia Rubisch, Claudio Michaelis, Matthias Bethge, Felix~A. Wichmann, and Wieland Brendel.
\newblock Imagenet-trained {CNN}s are biased towards texture; increasing shape bias improves accuracy and robustness.
\newblock In \emph{International Conference on Learning Representations}, 2019.

\bibitem[Gould et~al.(2024)Gould, Ong, Ogden, and Conmy]{gould2024successor}
Rhys Gould, Euan Ong, George Ogden, and Arthur Conmy.
\newblock Successor heads: Recurring, interpretable attention heads in the wild.
\newblock In \emph{The Twelfth International Conference on Learning Representations}, 2024.

\bibitem[Goyal and Bengio(2022)]{goyal2022inductive}
Anirudh Goyal and Yoshua Bengio.
\newblock Inductive biases for deep learning of higher-level cognition.
\newblock \emph{Proceedings of the Royal Society A}, 478\penalty0 (2266):\penalty0 20210068, 2022.

\bibitem[Graham and Poulin-Dubois(1999)]{graham1999infants}
Susan~A Graham and Diane Poulin-Dubois.
\newblock Infants' reliance on shape to generalize novel labels to animate and inanimate objects.
\newblock \emph{Journal of child language}, 26\penalty0 (2):\penalty0 295--320, 1999.

\bibitem[Gurnee and Tegmark()]{gurnee2023lan-space-time}
Wes Gurnee and Max Tegmark.
\newblock Language models represent space and time.
\newblock In \emph{The Twelfth International Conference on Learning Representations}.

\bibitem[Heinze-Deml and Bouchacourt(2020)]{heinze2020gscan-auxiliary}
Christina Heinze-Deml and Diane Bouchacourt.
\newblock Think before you act: A simple baseline for compositional generalization.
\newblock 2020.

\bibitem[Heinzerling and Inui(2024)]{heinzerling-inui-2024-monotonic}
Benjamin Heinzerling and Kentaro Inui.
\newblock Monotonic representation of numeric attributes in language models.
\newblock In \emph{Proceedings of the 62nd Annual Meeting of the Association for Computational Linguistics (Volume 2: Short Papers)}, pages 175--195, Bangkok, Thailand, 2024. Association for Computational Linguistics.

\bibitem[Higgins et~al.(2017)Higgins, Matthey, Pal, Burgess, Glorot, Botvinick, Mohamed, and Lerchner]{higgins2017betavae}
Irina Higgins, Loic Matthey, Arka Pal, Christopher Burgess, Xavier Glorot, Matthew Botvinick, Shakir Mohamed, and Alexander Lerchner.
\newblock beta-{VAE}: Learning basic visual concepts with a constrained variational framework.
\newblock In \emph{International Conference on Learning Representations}, 2017.

\bibitem[Higgins et~al.(2018)Higgins, Amos, Pfau, Racaniere, Matthey, Rezende, and Lerchner]{higgins2018definitiondisentangledrepresentations}
Irina Higgins, David Amos, David Pfau, Sebastien Racaniere, Loic Matthey, Danilo Rezende, and Alexander Lerchner.
\newblock Towards a definition of disentangled representations, 2018.

\bibitem[Ito et~al.(2022{\natexlab{a}})Ito, Klinger, Schultz, Murray, Cole, and Rigotti]{ito2022compositional}
Takuya Ito, Tim Klinger, Doug Schultz, John Murray, Michael Cole, and Mattia Rigotti.
\newblock Compositional generalization through abstract representations in human and artificial neural networks.
\newblock \emph{Advances in Neural Information Processing Systems}, 35:\penalty0 32225--32239, 2022{\natexlab{a}}.

\bibitem[Ito et~al.(2022{\natexlab{b}})Ito, Klinger, Schultz, Murray, Cole, and Rigotti]{ito2022parallelism-score}
Takuya Ito, Tim Klinger, Doug Schultz, John Murray, Michael Cole, and Mattia Rigotti.
\newblock Compositional generalization through abstract representations in human and artificial neural networks.
\newblock \emph{Advances in neural information processing systems}, 35:\penalty0 32225--32239, 2022{\natexlab{b}}.

\bibitem[Jiang et~al.(2022)Jiang, Zhou, and Bansal]{jiang-etal-2022-mutual}
Yichen Jiang, Xiang Zhou, and Mohit Bansal.
\newblock Mutual exclusivity training and primitive augmentation to induce compositionality.
\newblock In \emph{Proceedings of the 2022 Conference on Empirical Methods in Natural Language Processing}, pages 11778--11793, Abu Dhabi, United Arab Emirates, 2022. Association for Computational Linguistics.

\bibitem[Johnson et~al.(2017)Johnson, Hariharan, Van Der~Maaten, Fei-Fei, Lawrence~Zitnick, and Girshick]{johnson2017clevr}
Justin Johnson, Bharath Hariharan, Laurens Van Der~Maaten, Li Fei-Fei, C Lawrence~Zitnick, and Ross Girshick.
\newblock Clevr: A diagnostic dataset for compositional language and elementary visual reasoning.
\newblock In \emph{Proceedings of the IEEE conference on computer vision and pattern recognition}, pages 2901--2910, 2017.

\bibitem[Keysers et~al.(2019)Keysers, Sch{\"a}rli, Scales, Buisman, Furrer, Kashubin, Momchev, Sinopalnikov, Stafiniak, Tihon, et~al.]{keysers2019cfq}
Daniel Keysers, Nathanael Sch{\"a}rli, Nathan Scales, Hylke Buisman, Daniel Furrer, Sergii Kashubin, Nikola Momchev, Danila Sinopalnikov, Lukasz Stafiniak, Tibor Tihon, et~al.
\newblock Measuring compositional generalization: A comprehensive method on realistic data.
\newblock In \emph{International Conference on Learning Representations}, 2019.

\bibitem[Kim and Linzen(2020)]{kim2020cogs}
Najoung Kim and Tal Linzen.
\newblock Cogs: A compositional generalization challenge based on semantic interpretation.
\newblock In \emph{Proceedings of the 2020 Conference on Empirical Methods in Natural Language Processing (EMNLP)}, pages 9087--9105, 2020.

\bibitem[Kingma and Ba(2015)]{KingBa2015adam}
Diederik Kingma and Jimmy Ba.
\newblock Adam: A method for stochastic optimization.
\newblock In \emph{International Conference on Learning Representations (ICLR)}, San Diega, CA, USA, 2015.

\bibitem[Kuka{\v{c}}ka et~al.(2017)Kuka{\v{c}}ka, Golkov, and Cremers]{kukavcka2017regularization}
Jan Kuka{\v{c}}ka, Vladimir Golkov, and Daniel Cremers.
\newblock Regularization for deep learning: A taxonomy.
\newblock \emph{arXiv preprint arXiv:1710.10686}, 2017.

\bibitem[Kuo et~al.(2020)Kuo, Katz, and Barbu]{kuo2020gscan-recursive}
Yen-Ling Kuo, Boris Katz, and Andrei Barbu.
\newblock Compositional networks enable systematic generalization for grounded language understanding.
\newblock \emph{arXiv preprint arXiv:2008.02742}, 2020.

\bibitem[Lake and Baroni(2018)]{lake2017original-scan}
Brenden Lake and Marco Baroni.
\newblock Generalization without systematicity: On the compositional skills of sequence-to-sequence recurrent networks.
\newblock In \emph{International conference on machine learning}, pages 2873--2882. PMLR, 2018.

\bibitem[Lake(2019)]{lake2019meta-seq2seq}
Brenden~M Lake.
\newblock Compositional generalization through meta sequence-to-sequence learning.
\newblock In \emph{Advances in Neural Information Processing Systems 32}, pages 9791--9801. Curran Associates, Inc., 2019.

\bibitem[Lake and Baroni(2023)]{lake2023human}
Brenden~M Lake and Marco Baroni.
\newblock Human-like systematic generalization through a meta-learning neural network.
\newblock \emph{Nature}, 623\penalty0 (7985):\penalty0 115--121, 2023.

\bibitem[Lake et~al.(2017)Lake, Ullman, Tenenbaum, and Gershman]{lake2017building}
Brenden~M Lake, Tomer~D Ullman, Joshua~B Tenenbaum, and Samuel~J Gershman.
\newblock Building machines that learn and think like people.
\newblock \emph{Behavioral and brain sciences}, 40:\penalty0 e253, 2017.

\bibitem[Li et~al.(2022)Li, Li, Xiong, and Hoi]{li2022blip}
Junnan Li, Dongxu Li, Caiming Xiong, and Steven Hoi.
\newblock Blip: Bootstrapping language-image pre-training for unified vision-language understanding and generation.
\newblock In \emph{International conference on machine learning}, pages 12888--12900. PMLR, 2022.

\bibitem[Li et~al.(2023)Li, Li, Savarese, and Hoi]{li2023blip}
Junnan Li, Dongxu Li, Silvio Savarese, and Steven Hoi.
\newblock Blip-2: Bootstrapping language-image pre-training with frozen image encoders and large language models.
\newblock In \emph{International conference on machine learning}, pages 19730--19742. PMLR, 2023.

\bibitem[Li et~al.(2020)Li, Yin, Li, Zhang, Hu, Zhang, Wang, Hu, Dong, Wei, et~al.]{li2020oscar}
Xiujun Li, Xi Yin, Chunyuan Li, Pengchuan Zhang, Xiaowei Hu, Lei Zhang, Lijuan Wang, Houdong Hu, Li Dong, Furu Wei, et~al.
\newblock Oscar: Object-semantics aligned pre-training for vision-language tasks.
\newblock In \emph{Computer Vision--ECCV 2020: 16th European Conference, Glasgow, UK, August 23--28, 2020, Proceedings, Part XXX 16}, pages 121--137. Springer, 2020.

\bibitem[Liu et~al.(2020)Liu, An, Lou, Chen, Lin, Gao, Zhou, Zheng, and Zhang]{liu2020compositional}
Qian Liu, Shengnan An, Jian-Guang Lou, Bei Chen, Zeqi Lin, Yan Gao, Bin Zhou, Nanning Zheng, and Dongmei Zhang.
\newblock Compositional generalization by learning analytical expressions.
\newblock \emph{Advances in Neural Information Processing Systems}, 33:\penalty0 11416--11427, 2020.

\bibitem[Long et~al.(2015)Long, Shelhamer, and Darrell]{long2015fully}
Jonathan Long, Evan Shelhamer, and Trevor Darrell.
\newblock Fully convolutional networks for semantic segmentation.
\newblock In \emph{Proceedings of the IEEE conference on computer vision and pattern recognition}, pages 3431--3440, 2015.

\bibitem[Marcus(2003)]{marcus2003algebraic}
Gary~F Marcus.
\newblock \emph{The algebraic mind: Integrating connectionism and cognitive science}.
\newblock MIT press, 2003.

\bibitem[Marks and Tegmark(2024)]{marks2024geometry-truth}
Samuel Marks and Max Tegmark.
\newblock The geometry of truth: Emergent linear structure in large language model representations of true/false datasets.
\newblock In \emph{First Conference on Language Modeling}, 2024.

\bibitem[Mikolov et~al.(2013)Mikolov, Yih, and Zweig]{mikolov2013linguistic}
Tom{\'a}{\v{s}} Mikolov, Wen-tau Yih, and Geoffrey Zweig.
\newblock Linguistic regularities in continuous space word representations.
\newblock In \emph{Proceedings of the 2013 conference of the north american chapter of the association for computational linguistics: Human language technologies}, pages 746--751, 2013.

\bibitem[Mitchell(2021)]{mitchell2021abstraction}
Melanie Mitchell.
\newblock Abstraction and analogy-making in artificial intelligence.
\newblock \emph{Annals of the New York Academy of Sciences}, 1505\penalty0 (1):\penalty0 79--101, 2021.

\bibitem[Montero et~al.(2021)Montero, Ludwig, Costa, Malhotra, and Bowers]{montero2021the}
Milton~Llera Montero, Casimir~JH Ludwig, Rui~Ponte Costa, Gaurav Malhotra, and Jeffrey Bowers.
\newblock The role of disentanglement in generalisation.
\newblock In \emph{International Conference on Learning Representations}, 2021.

\bibitem[Nye et~al.(2020)Nye, Solar-Lezama, Tenenbaum, and Lake]{nye2020lcomp-program-synth}
Maxwell~I Nye, Armando Solar-Lezama, Joshua~B Tenenbaum, and Brenden~M Lake.
\newblock Learning compositional rules via neural program synthesis.
\newblock \emph{arXiv preprint arXiv:2003.05562}, 2020.

\bibitem[Olah et~al.(2020)Olah, Cammarata, Schubert, Goh, Petrov, and Carter]{olah2020zoom}
Chris Olah, Nick Cammarata, Ludwig Schubert, Gabriel Goh, Michael Petrov, and Shan Carter.
\newblock Zoom in: An introduction to circuits.
\newblock \emph{Distill}, 5\penalty0 (3):\penalty0 e00024--001, 2020.

\bibitem[Ontan{\'o}n et~al.(2021)Ontan{\'o}n, Ainslie, Cvicek, and Fisher]{ontanon2021transformer-sys-gen}
Santiago Ontan{\'o}n, Joshua Ainslie, Vaclav Cvicek, and Zachary Fisher.
\newblock Making transformers solve compositional tasks.
\newblock \emph{arXiv preprint arXiv:2108.04378}, 2021.

\bibitem[Park et~al.(2024)Park, Choe, and Veitch]{park2024linear}
Kiho Park, Yo~Joong Choe, and Victor Veitch.
\newblock The linear representation hypothesis and the geometry of large language models.
\newblock In \emph{Proceedings of the 41st International Conference on Machine Learning}, pages 39643--39666, 2024.

\bibitem[Paszke et~al.(2019)Paszke, Gross, Massa, Lerer, Bradbury, Chanan, Killeen, Lin, Gimelshein, Antiga, Desmaison, Kopf, Yang, DeVito, Raison, Tejani, Chilamkurthy, Steiner, Fang, Bai, and Chintala]{pytorch}
Adam Paszke, Sam Gross, Francisco Massa, Adam Lerer, James Bradbury, Gregory Chanan, Trevor Killeen, Zeming Lin, Natalia Gimelshein, Luca Antiga, Alban Desmaison, Andreas Kopf, Edward Yang, Zachary DeVito, Martin Raison, Alykhan Tejani, Sasank Chilamkurthy, Benoit Steiner, Lu Fang, Junjie Bai, and Soumith Chintala.
\newblock Pytorch: An imperative style, high-performance deep learning library.
\newblock In \emph{Advances in Neural Information Processing Systems 32}, pages 8024--8035. Curran Associates, Inc., 2019.

\bibitem[Patel et~al.(2022{\natexlab{a}})Patel, Bhattamishra, Blunsom, and Goyal]{patel-etal-2022-revisiting}
Arkil Patel, Satwik Bhattamishra, Phil Blunsom, and Navin Goyal.
\newblock Revisiting the compositional generalization abilities of neural sequence models.
\newblock In \emph{Proceedings of the 60th Annual Meeting of the Association for Computational Linguistics (Volume 2: Short Papers)}, pages 424--434, Dublin, Ireland, 2022{\natexlab{a}}. Association for Computational Linguistics.

\bibitem[Patel et~al.(2022{\natexlab{b}})Patel, Bhattamishra, Blunsom, and Goyal]{patel2022revisiting}
Arkil Patel, Satwik Bhattamishra, Phil Blunsom, and Navin Goyal.
\newblock Revisiting the compositional generalization abilities of neural sequence models.
\newblock In \emph{Proceedings of the 60th Annual Meeting of the Association for Computational Linguistics (Volume 2: Short Papers)}, pages 424--434, 2022{\natexlab{b}}.

\bibitem[Pearl(2009)]{causal_inference}
Judea Pearl.
\newblock \emph{Causality: Models, Reasoning and Inference}.
\newblock Cambridge University Press, USA, 2nd edition, 2009.

\bibitem[Peters et~al.(2016)Peters, Bühlmann, and Meinshausen]{Peters2016}
Jonas Peters, Peter Bühlmann, and Nicolai Meinshausen.
\newblock {Causal Inference by using Invariant Prediction: Identification and Confidence Intervals}.
\newblock \emph{Journal of the Royal Statistical Society Series B: Statistical Methodology}, 78\penalty0 (5):\penalty0 947--1012, 2016.

\bibitem[Prystawski et~al.(2024)Prystawski, Li, and Goodman]{prystawski2024think}
Ben Prystawski, Michael Li, and Noah Goodman.
\newblock Why think step by step? reasoning emerges from the locality of experience.
\newblock \emph{Advances in Neural Information Processing Systems}, 36, 2024.

\bibitem[Qiu et~al.(2021)Qiu, Hu, Zhang, Shaw, and Sha]{qiu2021gscan-solved}
Linlu Qiu, Hexiang Hu, Bowen Zhang, Peter Shaw, and Fei Sha.
\newblock Systematic generalization on gscan: What is nearly solved and what is next?
\newblock In \emph{Proceedings of the 2021 Conference on Empirical Methods in Natural Language Processing}, pages 2180--2188, 2021.

\bibitem[Qiu et~al.(2022)Qiu, Shaw, Pasupat, Nowak, Linzen, Sha, and Toutanova]{qiu2022improving}
Linlu Qiu, Peter Shaw, Panupong Pasupat, Pawel Nowak, Tal Linzen, Fei Sha, and Kristina Toutanova.
\newblock Improving compositional generalization with latent structure and data augmentation.
\newblock In \emph{Proceedings of the 2022 Conference of the North American Chapter of the Association for Computational Linguistics: Human Language Technologies}, pages 4341--4362, 2022.

\bibitem[Rahimi et~al.(2024)Rahimi, D'Amario, Yamada, Takemoto, Sasaki, and Boix]{rahimi2024d3}
Amir Rahimi, Vanessa D'Amario, Moyuru Yamada, Kentaro Takemoto, Tomotake Sasaki, and Xavier Boix.
\newblock D3: Data diversity design for systematic generalization in visual question answering.
\newblock \emph{Transactions on Machine Learning Research}, 2024.

\bibitem[Ravanbakhsh et~al.(2017)Ravanbakhsh, Schneider, and Poczos]{ravanbakhsh2017equivariance}
Siamak Ravanbakhsh, Jeff Schneider, and Barnabas Poczos.
\newblock Equivariance through parameter-sharing.
\newblock In \emph{International conference on machine learning}, pages 2892--2901. PMLR, 2017.

\bibitem[Ruis et~al.(2020)Ruis, Andreas, Baroni, Bouchacourt, and Lake]{ruis2020gscan}
Laura Ruis, Jacob Andreas, Marco Baroni, Diane Bouchacourt, and Brenden~M Lake.
\newblock A benchmark for systematic generalization in grounded language understanding.
\newblock \emph{Advances in Neural Information Processing Systems}, 33, 2020.

\bibitem[Satorras et~al.(2021)Satorras, Hoogeboom, and Welling]{satorras2021n}
V{\i}ctor~Garcia Satorras, Emiel Hoogeboom, and Max Welling.
\newblock E (n) equivariant graph neural networks.
\newblock In \emph{International conference on machine learning}, pages 9323--9332. PMLR, 2021.

\bibitem[Schott et~al.(2022)Schott, K{\"u}gelgen, Tr{\"a}uble, Gehler, Russell, Bethge, Sch{\"o}lkopf, Locatello, and Brendel]{schott2022visual}
Lukas Schott, Julius~Von K{\"u}gelgen, Frederik Tr{\"a}uble, Peter~Vincent Gehler, Chris Russell, Matthias Bethge, Bernhard Sch{\"o}lkopf, Francesco Locatello, and Wieland Brendel.
\newblock Visual representation learning does not generalize strongly within the same domain.
\newblock In \emph{International Conference on Learning Representations}, 2022.

\bibitem[Tigges et~al.(2024)Tigges, Hollinsworth, Geiger, and Nanda]{tigges2024language}
Curt Tigges, Oskar~John Hollinsworth, Atticus Geiger, and Neel Nanda.
\newblock Language models linearly represent sentiment.
\newblock In \emph{ICML 2024 Workshop on Mechanistic Interpretability}, 2024.

\bibitem[Vaswani et~al.(2017)Vaswani, Shazeer, Parmar, Uszkoreit, Jones, Gomez, Kaiser, and Polosukhin]{vaswani2017attention}
Ashish Vaswani, Noam Shazeer, Niki Parmar, Jakob Uszkoreit, Llion Jones, Aidan~N Gomez, {\L}ukasz Kaiser, and Illia Polosukhin.
\newblock Attention is all you need.
\newblock In \emph{Advances in neural information processing systems}, pages 5998--6008, 2017.

\bibitem[Wolpert et~al.(1995)Wolpert, Macready, et~al.]{wolpert1995no-free-lunch}
David~H Wolpert, William~G Macready, et~al.
\newblock No free lunch theorems for search.
\newblock Technical report, Citeseer, 1995.

\bibitem[Xu et~al.(2022)Xu, Niethammer, and Raffel]{xu2022compositional}
Zhenlin Xu, Marc Niethammer, and Colin~A Raffel.
\newblock Compositional generalization in unsupervised compositional representation learning: A study on disentanglement and emergent language.
\newblock \emph{Advances in Neural Information Processing Systems}, 35:\penalty0 25074--25087, 2022.

\bibitem[Yang et~al.(2022)Yang, Zhang, and Yang]{yang2022subs}
Jingfeng Yang, Le Zhang, and Diyi Yang.
\newblock Subs: Subtree substitution for compositional semantic parsing.
\newblock In \emph{Proceedings of the 2022 Conference of the North American Chapter of the Association for Computational Linguistics: Human Language Technologies}, pages 169--174, 2022.

\bibitem[Zhang et~al.(2024)Zhang, Zhao, Chen, Jiang, and Chen]{zhang2024feature-contamination}
Tianren Zhang, Chujie Zhao, Guanyu Chen, Yizhou Jiang, and Feng Chen.
\newblock Feature contamination: Neural networks learn uncorrelated features and fail to generalize.
\newblock In \emph{International Conference on Machine Learning}, pages 60446--60495. PMLR, 2024.

\bibitem[Zhou et~al.(2023)Zhou, Jiang, and Bansal]{zhou2023data-factors}
Xiang Zhou, Yichen Jiang, and Mohit Bansal.
\newblock Data factors for better compositional generalization.
\newblock In \emph{Proceedings of the 2023 Conference on Empirical Methods in Natural Language Processing}, pages 14549--14566, Singapore, 2023. Association for Computational Linguistics.

\end{thebibliography}
}
\clearpage
\setcounter{page}{1}
\maketitlesupplementary


\section{Hyper Parameters}

Images were resized to 224 × 224 pixels, and divided into patches of 16 × 16 pixels before being fed to the model. In the text, a masked language modeling (MLM) probability of 0.15 was used to randomly mask text tokens before feeding them to the model.

The model used during our experiments is a Transformer \cite{vaswani2017attention} with a hidden layer dimension of 256, with 4 transformer layers and 4 attention heads. Training was carried out using the Adam optimizer \citep{KingBa2015adam} with a learning rate of $1 \times 10^{-4}$, a batch size of 256 and for 1000 epochs.

\section{Dataset Samples}

Below we display samples composed of images and their corresponding textual descriptions from the datasets described in Section \ref{sec:dataset}:

\noindent
\textbf{Sample 1}

\begin{figure}[h]
    \centering
    \includegraphics[width=0.8\linewidth]{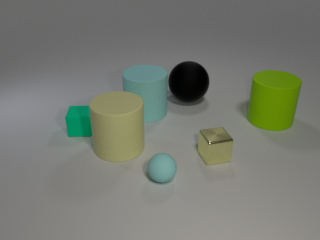}
\end{figure}

\noindent
\textbf{Textual Description}

\noindent
\texttt{\small small \#00ff80 rubber cube [SEP] large \#ffff80 rubber cylinder [SEP] large \#80ffff rubber cylinder [SEP] small \#ffff80 metal cube [SEP] large \#000000 rubber sphere [SEP] small \#80ffff rubber sphere [SEP] large \#80ff00 rubber cylinder
}

\vspace{1em}

\noindent
\textbf{Sample 2}

\begin{figure}[!h]
    \centering
    \includegraphics[width=0.8\linewidth]{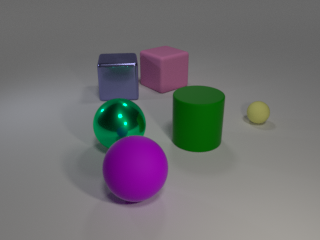}
\end{figure}

\noindent
\textbf{Textual Description}

\noindent
\texttt{\small large \#005500 rubber cylinder [SEP] large \#5555aa metal cube [SEP] large \#ff55aa rubber cube [SEP] small \#ffff55 rubber sphere [SEP] large \#aa00ff rubber sphere [SEP] large \#00ff55 metal sphere}

\vspace{1em}

\noindent
\textbf{Sample 3}

\begin{figure}[h]
    \centering
    \includegraphics[width=0.8\linewidth]{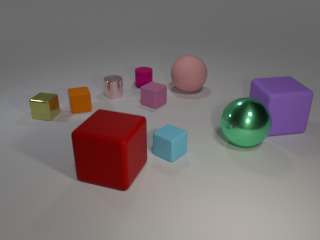}
\end{figure}

\noindent
\textbf{Textual Description}

\noindent
\texttt{\small small \#bf0040 rubber cylinder [SEP] large \#40ff80 metal sphere [SEP] large \#8040ff rubber cube [SEP] large \#800000 rubber cube [SEP] small \#bf4080 rubber cube [SEP] small \#ffbfbf metal cylinder [SEP] small \#ff4000 rubber cube [SEP] small \#40bfff rubber cube [SEP] small \#bfbf40 metal cube [SEP] large \#ff8080 rubber sphere
}

\section{Correlation between P-Score and OOD Performance}

In Table \ref{tab:correlation_pshape} we show the Pearson correlation between different p-scores obtained for a model and their OOD performance on the shape task. Parallelism in the representations seems to strongly correlate with better OOD performance, suggesting that parallelism may be the property that is inducing SG in the models.

\begin{table}[h]
    \centering
    \begin{tabular}{lcccc}
        \toprule
        \textsc{Metric} & \textsc{Avg} & \textsc{DIV} & \textsc{BURST} & \textsc{LI} \\
        \toprule
        \textsc{Corr} & $0.73$ & $0.76$ & $0.69$ & $0.73$ \\
        \textsc{p-value} & $1.88e{-9}$ &  $7.17e{-4}$& $5.5e{-3}$ & $3.45e{-5}$ \\
        \bottomrule
    \end{tabular}
    \caption{Pearson Correlation between P-Scores and OOD performance on the \textit{shape} task broken down per training data property.  DIV: Diversity, BURST: Burstiness, LI: Latent Intervention.}
    \label{tab:correlation_pshape}
\end{table}

\newpage

\section{Detailed Experimental Results}

In tables \ref{tab:fig:base-attributes-performance}, \ref{tab:fig:shapes_subset}, \ref{tab:acc_by_colors}, \ref{tab:burstiness}, \ref{tab:latent_intervention}, and \ref{tab:smaller-models}, we exposed the values obtained that were used to generate most of the figures in the work above.
Table \ref{tab:fig:base-attributes-performance} shows the data used for Figure \ref{fig:base-attributes-performance}.
Table \ref{fig:shapes_subset} for Figure \ref{fig:shapes_subset}.
Table \ref{tab:acc_by_colors} for Figure \ref{fig:acc_by_colors}.
Table \ref{tab:burstiness} for Figure \ref{fig:latent_intervention}.
Table \ref{tab:latent_intervention} for Figure \ref{fig:burstiness}.
And, Table \ref{tab:smaller-models} for Figure \ref{fig:smaller-models}.

\begin{table}[h]
    \centering

\resizebox{0.95\linewidth}{!}{%
    \begin{tabular}{lccc}
    \toprule
     & Train & Test-ID & Test-OOD \\
    \midrule
    Color & 94.3 ± 0.0\% & 94.1 ± 0.1\% & 50.5 ± 1.1\% \\
    Shapes & 100.0 ± 0.0\% & 99.6 ± 0.0\% & 0.6 ± 0.1\% \\
    Materials & 98.0 ± 0.0\% & 96.9 ± 0.0\% & 87.8 ± 0.9\% \\
    Size & 98.1 ± 0.0\% & 98.0 ± 0.0\% & 91.2 ± 1.0\% \\
    \bottomrule
    \end{tabular}
}
    \caption{Table displaying values to form Figure \ref{fig:base-attributes-performance}. Accuracy for predicting different properties for different data splits for the baseline model (8 colors). ID performance for all tasks remains high, however OOD performance plummets for shape and color, suggesting the model is learning combinations of shape-color as features, instead of achieving SG. Unexpectedly, material and size also show a drop in OOD performance, even though the model has been exposed to all combinations of these attributes in training.}
    \label{tab:fig:base-attributes-performance}
\end{table}

\begin{table}[t]
\vspace{-38em}
    \centering

\resizebox{.95\linewidth}{!}{%
\begin{tabular}{m{0.8cm}cccc}
\toprule
 & \multicolumn{2}{c}{Color} & \multicolumn{2}{c}{Shape} \\
 Train Fract. & Test-ID & Test-OOD & Test-ID & Test-OOD \\
 
\midrule
\multicolumn{2}{l}{8 Colors} & & & \\
\midrule
1/8 & 91.5 ± 0.1\% & 47.4 ± 0.4\% & 97.1 ± 0.1\% & 2.1 ± 0.1\% \\
1/4 & 93.1 ± 0.1\% & 47.4 ± 0.5\% & 99.0 ± 0.0\% & 1.2 ± 0.1\% \\
1/2 & 93.9 ± 0.0\% & 47.3 ± 0.6\% & 99.2 ± 0.0\% & 0.6 ± 0.1\% \\
1/1 & 94.1 ± 0.1\% & 50.5 ± 1.1\% & 99.6 ± 0.0\% & 0.6 ± 0.1\% \\

\midrule
\multicolumn{2}{l}{216 Colors} & & & \\
\midrule
1/8 & 22.8 ± 2.4\% & 5.0 ± 0.9\% & 45.1 ± 0.6\% & 43.3 ± 0.5\% \\
1/4 & 79.6 ± 0.6\% & 60.4 ± 1.3\% & 89.1 ± 0.5\% & 81.0 ± 0.8\% \\
1/2 & 86.1 ± 0.0\% & 74.1 ± 0.4\% & 94.4 ± 0.1\% & 88.4 ± 0.2\% \\
1/1 & 88.1 ± 0.1\% & 77.9 ± 0.4\% & 96.3 ± 0.0\% & 90.0 ± 0.4\% \\
\bottomrule

\end{tabular}
}
    \caption{Table displaying values to form Figure \ref{fig:shapes_subset}. Accuracy for the \textit{shape} task for different amounts of training data for the baseline model (8 colors) vs model trained on 216 colors. (a) Increasing dataset size does not increase OOD performance but rather degrades it slightly.  (b) With increased \textit{diversity} much stronger OOD generalization is achieved, with models trained on only a quarter of the data severely outperforming the 8-color baseline.}
    \label{tab:fig:shapes_subset}
\end{table}

\begin{table*}[hbt!]
    \centering
    
\resizebox{\textwidth}{!}{%
\begin{tabular}{m{1cm}cccccccc}
\toprule
 & \multicolumn{2}{c}{Color} & \multicolumn{2}{c}{Shape}
 & \multicolumn{2}{c}{Size} & \multicolumn{2}{c}{Material} \\
 Num. Colors & Test-ID & Test-OOD & Test-ID & Test-OOD & Test-ID & Test-OOD & Test-ID & Test-OOD \\
\midrule
8 & 94.1 ± 0.1\% & 50.5 ± 1.1\% & 99.6 ± 0.0\% & 0.6 ± 0.1\% & 98.0 ± 0.0\% & 91.2 ± 1.0\% & 96.9 ± 0.0\% & 87.8 ± 0.9\% \\
27 & 91.7 ± 0.0\% & 67.5 ± 0.7\% & 96.9 ± 0.1\% & 1.5 ± 0.1\% & 99.5 ± 0.0\% & 97.6 ± 0.0\% & 98.6 ± 0.0\% & 96.7 ± 0.1\% \\
64 & 90.8 ± 0.1\% & 73.0 ± 0.5\% & 96.9 ± 0.0\% & 48.5 ± 0.6\% & 99.7 ± 0.0\% & 97.6 ± 0.0\% & 99.2 ± 0.0\% & 96.8 ± 0.0\% \\
125 & 89.0 ± 0.1\% & 67.0 ± 0.3\% & 96.1 ± 0.1\% & 81.8 ± 0.6\% & 99.8 ± 0.0\% & 97.8 ± 0.0\% & 99.2 ± 0.0\% & 96.9 ± 0.0\% \\
216 & 88.1 ± 0.1\% & 77.9 ± 0.4\% & 96.3 ± 0.0\% & 90.0 ± 0.4\% & 99.8 ± 0.0\% & 97.7 ± 0.0\% & 99.3 ± 0.0\% & 97.2 ± 0.0\% \\
\bottomrule
\end{tabular}
}
    \caption{Table displaying values to form Figure \ref{fig:acc_by_colors}. Accuracy versus the number of colors in $\mathcal{D}_{train}$ for $\mathcal{D}_{test-ID}$ and $\mathcal{D}_{test-OOD}$ for all tasks. Performance for the \textit{shape} task increases drastically for the OOD split as we increase colors, increasing 86\% in absolute terms over the 8-color baseline. Moreover, performance in the color task also tends to increase in the OOD split, while ID only suffers slightly, even though the task becomes significantly harder. Remarkably, the \textit{material} and \textit{size} task rapidly increase their $\mathcal{D}_{test-OOD}$ performance as color increases.}
    \label{tab:acc_by_colors}
\end{table*}

\begin{table*}[b]
    \centering

\resizebox{\textwidth}{!}{%
\begin{tabular}{m{1cm}m{1cm}cccccccc}
\toprule
 & & \multicolumn{2}{c}{Color} & \multicolumn{2}{c}{Shape}
 & \multicolumn{2}{c}{Size} & \multicolumn{2}{c}{Material} \\
 Num. Colors & P Bursty& Test-ID & Test-OOD & Test-ID & Test-OOD & Test-ID & Test-OOD & Test-ID & Test-OOD \\
\midrule
\multirow{3}{4em}{8} & 0.0 & 94.1 ± 0.1\% & 50.5 ± 1.1\% & 99.6 ± 0.0\% & 0.6 ± 0.1\% & 98.0 ± 0.0\% & 91.2 ± 1.0\% & 96.9 ± 0.0\% & 87.8 ± 0.9\% \\
 & 0.5 & 94.2 ± 0.0\% & 48.9 ± 0.9\% & 99.6 ± 0.0\% & 0.9 ± 0.2\% & 99.5 ± 0.0\% & 97.6 ± 0.0\% & 98.6 ± 0.0\% & 96.7 ± 0.1\% \\
 & 1.0 & 93.8 ± 0.0\% & 47.3 ± 0.1\% & 99.4 ± 0.0\% & 0.5 ± 0.0\% & 99.7 ± 0.0\% & 97.6 ± 0.0\% & 99.2 ± 0.0\% & 96.8 ± 0.0\% \\
\midrule
\multirow{3}{4em}{27} & 0.0 & 91.7 ± 0.0\% & 67.5 ± 0.7\% & 96.9 ± 0.1\% & 1.5 ± 0.1\% & 98.0 ± 0.0\% & 91.2 ± 1.0\% & 96.9 ± 0.0\% & 87.8 ± 0.9\% \\
 & 0.5 & 91.7 ± 0.0\% & 71.9 ± 0.2\% & 97.1 ± 0.0\% & 1.8 ± 0.3\% & 99.5 ± 0.0\% & 97.6 ± 0.0\% & 98.6 ± 0.0\% & 96.7 ± 0.1\% \\
 & 1.0 & 90.3 ± 0.0\% & 65.2 ± 0.8\% & 96.9 ± 0.0\% & 4.7 ± 0.4\% & 99.7 ± 0.0\% & 97.6 ± 0.0\% & 99.2 ± 0.0\% & 96.8 ± 0.0\% \\
\midrule
\multirow{3}{4em}{64} & 0.0 & 90.8 ± 0.1\% & 73.0 ± 0.5\% & 96.9 ± 0.0\% & 48.5 ± 0.6\% & 98.0 ± 0.0\% & 91.2 ± 1.0\% & 96.9 ± 0.0\% & 87.8 ± 0.9\% \\
& 0.5 & 89.8 ± 0.5\% & 73.8 ± 1.0\% & 97.0 ± 0.1\% & 60.5 ± 2.0\% & 99.5 ± 0.0\% & 97.6 ± 0.0\% & 98.6 ± 0.0\% & 96.7 ± 0.1\% \\
& 1.0 & 83.5 ± 0.2\% & 58.7 ± 1.0\% & 96.9 ± 0.1\% & 63.3 ± 0.9\% & 99.7 ± 0.0\% & 97.6 ± 0.0\% & 99.2 ± 0.0\% & 96.8 ± 0.0\% \\
\midrule
\multirow{3}{4em}{125} & 0.0 & 89.0 ± 0.1\% & 67.0 ± 0.3\% & 96.1 ± 0.1\% & 81.8 ± 0.6\% & 98.0 ± 0.0\% & 91.2 ± 1.0\% & 96.9 ± 0.0\% & 87.8 ± 0.9\% \\
& 0.5 & 88.9 ± 0.1\% & 68.5 ± 0.3\% & 96.8 ± 0.0\% & 80.6 ± 1.2\% & 99.5 ± 0.0\% & 97.6 ± 0.0\% & 98.6 ± 0.0\% & 96.7 ± 0.1\% \\
& 1.0 & 80.7 ± 0.2\% & 64.6 ± 0.3\% & 96.8 ± 0.0\% & 79.4 ± 0.5\% & 99.7 ± 0.0\% & 97.6 ± 0.0\% & 99.2 ± 0.0\% & 96.8 ± 0.0\% \\
\midrule
\multirow{3}{4em}{216} & 0.0 & 88.1 ± 0.1\% & 77.9 ± 0.4\% & 96.3 ± 0.0\% & 90.0 ± 0.4\% & 98.0 ± 0.0\% & 91.2 ± 1.0\% & 96.9 ± 0.0\% & 87.8 ± 0.9\% \\
& 0.5 & 87.3 ± 0.0\% & 77.7 ± 0.2\% & 96.7 ± 0.0\% & 91.4 ± 0.2\% & 99.5 ± 0.0\% & 97.6 ± 0.0\% & 98.6 ± 0.0\% & 96.7 ± 0.1\% \\
& 1.0 & 71.9 ± 0.2\% & 57.3 ± 0.4\% & 96.7 ± 0.0\% & 92.4 ± 0.3\% & 99.7 ± 0.0\% & 97.6 ± 0.0\% & 99.2 ± 0.0\% & 96.8 ± 0.0\% \\
\bottomrule
\end{tabular}
}  
    \caption{Table displaying values to form Figure \ref{fig:burstiness}. Accuracy fo all tasks in test-ID and test-OOD for different levels of \textit{burstiness} over \textit{color} for various numbers of colors. Limiting the number of colors available for each image during training allows the model to gain up to 14.8\% more accuracy over the baseline. The \textit{color} task, however, suffers up to 14.3\% decline as the \textit{color} task becomes easier to memorize.}
    \label{tab:burstiness}
\end{table*}

\begin{table*}[t]
    \centering

\resizebox{\textwidth}{!}{%
\begin{tabular}{m{1cm}m{1cm}cccccccc}
\toprule
 & & \multicolumn{2}{c}{Color} & \multicolumn{2}{c}{Shape}
 & \multicolumn{2}{c}{Size} & \multicolumn{2}{c}{Material} \\
 Num. Colors & Jitter & Test-ID & Test-OOD & Test-ID & Test-OOD & Test-ID & Test-OOD & Test-ID & Test-OOD \\
\midrule
\multirow{3}{4em}{8} & 0.00 & 94.1 ± 0.1\% & 50.5 ± 1.1\% & 99.6 ± 0.0\% & 0.6 ± 0.1\% & 98.0 ± 0.0\% & 91.2 ± 1.0\% & 96.9 ± 0.0\% & 87.8 ± 0.9\% \\
 & 0.05 & 94.1 ± 0.1\% & 46.6 ± 0.4\% & 99.6 ± 0.0\% & 0.4 ± 0.0\% & 99.5 ± 0.0\% & 97.6 ± 0.0\% & 98.6 ± 0.0\% & 96.7 ± 0.1\% \\
 & 0.10 & 94.1 ± 0.1\% & 48.0 ± 0.7\% & 99.6 ± 0.0\% & 0.9 ± 0.2\% & 99.7 ± 0.0\% & 97.6 ± 0.0\% & 99.2 ± 0.0\% & 96.8 ± 0.0\% \\
 & 0.50 & 93.9 ± 0.0\% & 47.1 ± 0.4\% & 99.7 ± 0.0\% & 0.3 ± 0.1\% & 99.8 ± 0.0\% & 97.8 ± 0.0\% & 99.2 ± 0.0\% & 96.9 ± 0.0\% \\
\midrule
\multirow{3}{4em}{27} & 0.00 & 91.7 ± 0.0\% & 67.5 ± 0.7\% & 96.9 ± 0.1\% & 1.5 ± 0.1\% & 98.0 ± 0.0\% & 91.2 ± 1.0\% & 96.9 ± 0.0\% & 87.8 ± 0.9\% \\
 & 0.05 & 91.9 ± 0.0\% & 69.4 ± 1.5\% & 97.4 ± 0.0\% & 1.4 ± 0.2\% & 99.5 ± 0.0\% & 97.6 ± 0.0\% & 98.6 ± 0.0\% & 96.7 ± 0.1\% \\
 & 0.10 & 91.8 ± 0.1\% & 68.4 ± 1.0\% & 97.3 ± 0.1\% & 1.7 ± 0.2\% & 99.7 ± 0.0\% & 97.6 ± 0.0\% & 99.2 ± 0.0\% & 96.8 ± 0.0\% \\
 & 0.50 & 91.9 ± 0.0\% & 66.1 ± 2.0\% & 97.2 ± 0.0\% & 1.1 ± 0.2\% & 99.8 ± 0.0\% & 97.8 ± 0.0\% & 99.2 ± 0.0\% & 96.9 ± 0.0\% \\
\midrule
\multirow{3}{4em}{64} & 0.00 & 90.8 ± 0.1\% & 73.0 ± 0.5\% & 96.9 ± 0.0\% & 48.5 ± 0.6\% & 98.0 ± 0.0\% & 91.2 ± 1.0\% & 96.9 ± 0.0\% & 87.8 ± 0.9\% \\
 & 0.05 & 90.9 ± 0.0\% & 68.5 ± 0.6\% & 97.1 ± 0.0\% & 57.1 ± 0.1\% & 99.5 ± 0.0\% & 97.6 ± 0.0\% & 98.6 ± 0.0\% & 96.7 ± 0.1\% \\
 & 0.10 & 91.0 ± 0.0\% & 67.1 ± 0.3\% & 97.3 ± 0.0\% & 62.3 ± 3.0\% & 99.7 ± 0.0\% & 97.6 ± 0.0\% & 99.2 ± 0.0\% & 96.8 ± 0.0\% \\
 & 0.50 & 91.1 ± 0.0\% & 68.6 ± 0.6\% & 97.8 ± 0.0\% & 63.8 ± 0.9\% & 99.8 ± 0.0\% & 97.8 ± 0.0\% & 99.2 ± 0.0\% & 96.9 ± 0.0\% \\
\midrule
\multirow{3}{4em}{125} & 0.00 & 89.0 ± 0.1\% & 67.0 ± 0.3\% & 96.1 ± 0.1\% & 81.8 ± 0.6\% & 98.0 ± 0.0\% & 91.2 ± 1.0\% & 96.9 ± 0.0\% & 87.8 ± 0.9\% \\
 & 0.05 & 89.8 ± 0.0\% & 71.1 ± 0.4\% & 96.8 ± 0.1\% & 85.0 ± 0.5\% & 99.5 ± 0.0\% & 97.6 ± 0.0\% & 98.6 ± 0.0\% & 96.7 ± 0.1\% \\
 & 0.10 & 89.7 ± 0.1\% & 68.4 ± 0.5\% & 96.7 ± 0.1\% & 79.2 ± 0.7\% & 99.7 ± 0.0\% & 97.6 ± 0.0\% & 99.2 ± 0.0\% & 96.8 ± 0.0\% \\
 & 0.50 & 89.8 ± 0.1\% & 70.6 ± 1.1\% & 96.9 ± 0.0\% & 85.8 ± 0.9\% & 99.8 ± 0.0\% & 97.8 ± 0.0\% & 99.2 ± 0.0\% & 96.9 ± 0.0\% \\
\midrule
\multirow{3}{4em}{216} & 0.00 & 88.1 ± 0.1\% & 77.9 ± 0.4\% & 96.3 ± 0.0\% & 90.0 ± 0.4\% & 98.0 ± 0.0\% & 91.2 ± 1.0\% & 96.9 ± 0.0\% & 87.8 ± 0.9\% \\
 & 0.05 & 88.4 ± 0.0\% & 79.8 ± 0.3\% & 96.8 ± 0.1\% & 92.2 ± 0.1\% & 99.5 ± 0.0\% & 97.6 ± 0.0\% & 98.6 ± 0.0\% & 96.7 ± 0.1\% \\
 & 0.10 & 88.0 ± 0.1\% & 75.6 ± 0.2\% & 96.4 ± 0.1\% & 91.0 ± 0.3\% & 99.7 ± 0.0\% & 97.6 ± 0.0\% & 99.2 ± 0.0\% & 96.8 ± 0.0\% \\
 & 0.50 & 88.4 ± 0.1\% & 78.7 ± 0.7\% & 96.7 ± 0.1\% & 92.0 ± 0.1\% & 99.8 ± 0.0\% & 97.8 ± 0.0\% & 99.2 ± 0.0\% & 96.9 ± 0.0\% \\
\bottomrule
\end{tabular}
}    
    \caption{Table displaying values to form Figure \ref{fig:latent_intervention}. Accuracy after applying latent intervention for all tasks in test-ID and test-OOD for different levels of latent intervention of the \textit{color} latent attribute for various numbers of colors. Altering the color hue randomly during training allows the model to gain up to 15\% more OOD accuracy over the baseline.}
    \label{tab:latent_intervention}
\end{table*}

\begin{table*}[h]
    \centering

\resizebox{\textwidth}{!}{%
\begin{tabular}{m{1cm}m{1cm}cccccccc}
\toprule
 & & \multicolumn{2}{c}{Color} & \multicolumn{2}{c}{Shape}
 & \multicolumn{2}{c}{Size} & \multicolumn{2}{c}{Material} \\
 Num. Colors & Hidden Size & Test-ID & Test-OOD & Test-ID & Test-OOD & Test-ID & Test-OOD & Test-ID & Test-OOD \\
\midrule
\multirow{4}{4em}{8} & 32 & 79.3 ± 0.1\% & 4.6 ± 0.6\% & 90.8 ± 0.8\% & 0.0 ± 0.0\% & 98.0 ± 0.0\% & 91.2 ± 1.0\% & 96.9 ± 0.0\% & 87.8 ± 0.9\% \\
& 64 & 87.0 ± 1.6\% & 18.1 ± 1.9\% & 98.1 ± 0.4\% & 0.0 ± 0.0\% & 99.5 ± 0.0\% & 97.6 ± 0.0\% & 98.6 ± 0.0\% & 96.7 ± 0.1\% \\
& 128 & 93.8 ± 0.1\% & 43.9 ± 0.7\% & 99.5 ± 0.0\% & 0.2 ± 0.0\% & 99.7 ± 0.0\% & 97.6 ± 0.0\% & 99.2 ± 0.0\% & 96.8 ± 0.0\% \\
& 256 & 94.1 ± 0.1\% & 50.5 ± 1.1\% & 99.6 ± 0.0\% & 0.6 ± 0.1\% & 99.8 ± 0.0\% & 97.8 ± 0.0\% & 99.2 ± 0.0\% & 96.9 ± 0.0\% \\
\midrule
\multirow{3}{4em}{216} & 256 & 88.1 ± 0.1\% & 77.9 ± 0.4\% & 96.3 ± 0.0\% & 90.0 ± 0.4\% & 98.0 ± 0.0\% & 91.2 ± 1.0\% & 96.9 ± 0.0\% & 87.8 ± 0.9\% \\
& 512 & 88.8 ± 0.0\% & 81.6 ± 0.2\% & 97.3 ± 0.0\% & 93.5 ± 0.1\% & 99.5 ± 0.0\% & 97.6 ± 0.0\% & 98.6 ± 0.0\% & 96.7 ± 0.1\% \\
& 1024 & 89.1 ± 0.1\% & 82.2 ± 0.1\% & 97.6 ± 0.0\% & 92.4 ± 0.2\% & 99.7 ± 0.0\% & 97.6 ± 0.0\% & 99.2 ± 0.0\% & 96.8 ± 0.0\% \\
\bottomrule
\end{tabular}
}
    \caption{Table displaying values to form Figure \ref{fig:smaller-models}. ID and OOD accuracy for models trained with different values for their hidden dimensions on a training set with 8 and 216 colors for all tasks.}
    \label{tab:smaller-models}
\end{table*}






\end{document}